\documentclass[journal]{IEEEtran}
\usepackage{amsmath}
\usepackage[colorlinks,
linkcolor=black,
anchorcolor=black,
citecolor=black]{hyperref}
\usepackage{booktabs}
\usepackage{multirow}
\usepackage{makecell}

\usepackage{autobreak}
\usepackage{color}
\usepackage{amssymb}
\ifCLASSINFOpdf
\usepackage[pdftex]{graphicx}
\graphicspath{{../pdf/}{../jpeg/}}
\DeclareGraphicsExtensions{.pdf,.jpeg,.png}
\else
\fi
\begin{document}
\title{Exploring Modality-shared Appearance Features and Modality-invariant Relation Features for Cross-modality Person Re-Identification}
\author{Nianchang Huang,~Jianan Liu,~Qiang Zhang*,~Jungong Han*% <-this % stops a space
	\thanks{Nianchang Huang, Jianan Liu, Qiang Zhang are with Center for Complex Systems, School of Mechano-Electronic Engineering, Xidian University, Xi’an, Shaanxi 710071, China. Email: nchuang@stu.xidian.edu.cn, jianan\_liu@stu.xidian.edu.cn and qzhang@xidian.edu.cn.}% <-this % stops a space
	\thanks{Jungong Han is with Computer Science Department, Aberystwyth University, SY23 3FL, UK. Email: jungonghan77@gmail.com }
	\thanks{*Corresponding authors: Qiang Zhang and Jungong Han.} }	

%
%
%
%
%

% and the end of the author line. \emph{i.e.,} if you had this:

\markboth{Journal of IEEE Transactions on Image Processing}%
{Shell \MakeLowercase{\textit{et al.}}: Exploring Modality-shared Appearance Features and Modality-invariant Relation Features for Cross-modality Person Re-Identification}



\maketitle

\begin{abstract}

Most existing cross-modality person re-identification works rely on discriminative modality-shared features for reducing cross-modality variations and intra-modality variations. Despite some initial success, such modality-shared appearance features cannot capture enough modality-invariant discriminative information due to a massive discrepancy between RGB and infrared images.  To address this issue, on the top of appearance features, we further capture the modality-invariant relations among different person parts (referred to as modality-invariant relation features), which are the complement to those modality-shared appearance features and help to identify persons with similar appearances but different body shapes. To this end, a Multi-level Two-streamed Modality-shared Feature Extraction (MTMFE) sub-network is designed, where the modality-shared appearance features and modality-invariant relation features are first extracted in a shared 2D feature space and a shared 3D feature space, respectively. The two features are then fused into the final modality-shared features such that both cross-modality variations and intra-modality variations can be reduced.   Besides, a novel cross-modality quadruplet loss is proposed to further reduce the cross-modality variations.  Experimental results on several benchmark datasets demonstrate that our proposed method exceeds state-of-the-art algorithms by a noticeable margin. 
\end{abstract}


% \fi
\IEEEpeerreviewmaketitle

\section{Introduction}\label{sec::intro}

\IEEEPARstart{P}{erson} Re-IDentification (Re-ID) aims to match a given pedestrian from disjointed camera views, which plays an important role in intelligent video surveillance \cite{r18} and people tracking \cite{r19}.  Recently, visible image based person Re-ID models (\emph{i.e.,} RGB-RGB images matching) have attracted wide attention and made great progress \cite{r36,r37,r38,r41,r42,r43}. However, visible cameras are sensitive to light conditions, which cannot capture informative images under inadequate illumination (\emph{e.g.,} at night). In such cases, the visible image based person Re-ID methods may show a dramatic performance degradation. Compared with visible cameras, infrared cameras are less dependent on light conditions and may capture more informative infrared (IR) images in challenging  illuminations. Considering that, to capture extensive information, many surveillance cameras support an automatic switch from the visible camera to the infrared camera when the light condition is worse \cite{r1}. Accordingly, cross-modality person Re-ID (\emph{i.e.,} RGB-IR image matching) has received increasing interests recently \cite{r1, r2,r3,r4,r5}. This further promotes the real-world applications of person Re-ID. 

\begin{figure}[!t]
	\centering
	\includegraphics[width=0.45\textwidth]{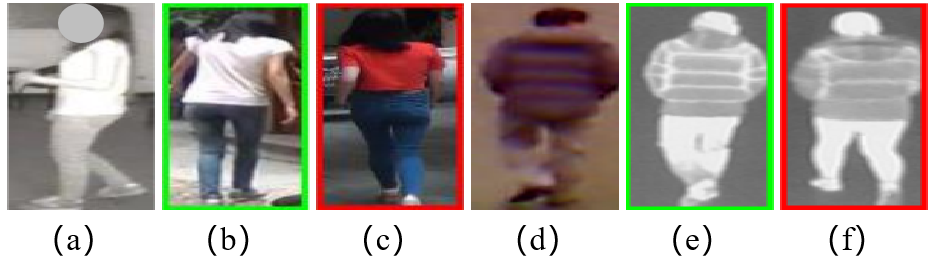}
	\caption{Examples of cross-modality person Re-ID. (a) Query of IR images. (b) and (c) Matched RGB images corresponding to (a). (d) Query of RGB images. (e) and (f) Matched IR images corresponding to (d). Images marked by green boxes denote the right matches, while those marked by red boxes denote the wrong matches. }
	\label{fig0}
\end{figure}

As shown in Fig. \ref{fig0}, the difficulties of cross-modality person Re-ID mainly lie in the following two aspects, \emph{i.e.,} cross-modality variations and intra-modality variations \cite{r5,r6, r7, r9}. Cross-modality variations are caused by the modality differences between visible (RGB) and infrared (IR) images, which may lead to diverse distributions of single-modality RGB and IR features. Intra-modality variations may result from many factors, \emph{e.g.,} different viewpoints, human poses changing and self-occlusions, which further bring difficulties in cross-modality person Re-ID. Most existing cross-modality person Re-ID models \cite{r17,r16,r13,r12} follow the idea of modality-shared feature learning, which aims at extracting discriminative modality-shared features (or modality-invariant features) to simultaneously reduce cross-modality variations and intra-modality variations. Concretely, these models first separately extract modality-specific features from the input RGB and IR images. Then, the modality-specific features are projected into a shared feature space, on which modality-shared features are extracted for cross-modality person Re-ID. 

\begin{figure}[!t]
	\centering
	\includegraphics[width=0.45\textwidth]{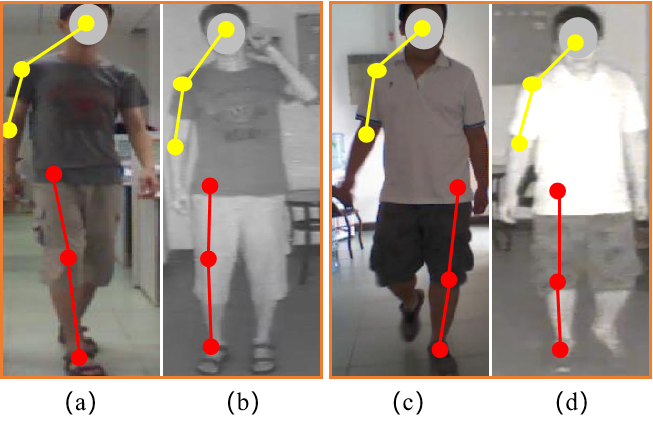}
	\caption{Illustrations of modality-invariant relations of different human body parts. (a) and (b) The RGB and IR images of a person. (c) and (d) The RGB and IR images of another person.  The yellow lines mean the ratios of the distance of the head to the shoulder with that of the shoulder to the elbow, while the red lines mean the ratio of the distance of the hip to the knee with that of the knee to the foot. By virtue of these modality-invariant relations, the persons in the images of different modalities (\emph{e.g.,} (a) and (d), (b) and (c), respectively)  can be identified as different ones.}
	\label{fig0_2}
\end{figure}

\begin{figure}[!t]
	\centering
	\includegraphics[width=0.45\textwidth]{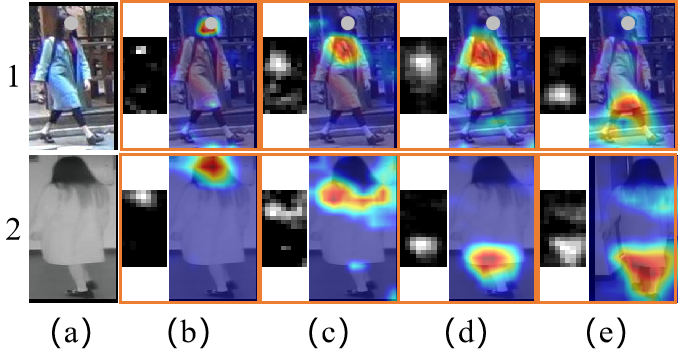}
	\caption{Visualization of some appearance features. (a) Input RGB or IR image. (b)-(e) The appearance features and corresponding heat maps in input images. The heat maps are obtained by mapping those appearance features into input images, which reflect the regions of corresponding features' focus. It can be seen that different channels of appearance features mainly focus on different person parts.}
	\label{fig0_1}
\end{figure}

Similar to single-modality person Re-ID \cite{r86}, most existing cross-modality person Re-ID models extract appearance features from the shared feature space of input RGB and IR images as the final modality-shared features. However, the appearances of the same person vary significantly from RGB images to IR images due to the large modality discrepancies caused by different imaging mechanisms. As a result, only extracting modality-shared appearance features may not capture enough discriminative information for identifying different persons in cross-modality person Re-ID. For examples, Fig. \ref{fig0}(a) and (b) are the same identity in the RGB and IR images, respectively, but Fig. \ref{fig0}(c) is another identity in an RGB image. Unfortunately, Fig. \ref{fig0}(c) is still wrongly identified as the same person as in Fig. \ref{fig0}(a) due to the large modality differences between RGB and IR images. Similarly, Fig. \ref{fig0}(f) is also wrongly matched as the person in Fig. \ref{fig0}(d).

%Accordingly, extracting discriminative modality-shared features for reducing cross-modality variations and intra-modality variations is one of the most important core issues for modality-shared feature learning based models. As in single-modality person Re-ID \cite{r86}, most existing cross-modality person Re-ID models extract modality-shared appearance features from the input RGB and IR images by using a shared sub-network. However, only using modality-shared appearance features may not capture enough discriminative information for identifying different persons in cross-modality person Re-ID due to the large modality differences between RGB and IR images. For examples, Fig. \ref{fig0}(a) and (b) are same identity and Fig. \ref{fig0}(c) is another one. However, Fig. \ref{fig0}(a) and (b) has different appearances with Fig. \ref{fig0}(b) or (c), due to the large modality differences. As a result, although Fig. \ref{fig0}(b) and (c) can be easily recognized as different persons in RGB images, Fig. \ref{fig0}(a), (b) and (c)  may be still recognized as same person in cross-modality person Re-ID.

Although those appearance features may be different in the images of different modalities, the relations among different person parts are invariant to modalities, which may help to identify persons with different body shapes, especially when their appearance features are similar in cross-modality person Re-ID.  For examples, as shown in Fig. \ref{fig0_2}, the two persons in the images of different modalities (\emph{i.e.,} persons in Fig. \ref{fig0_2}(a) and (d), respectively) can be easily identified as different ones via the ratios of the distance of the hip to the knee with that of the knee to the foot. Differently, the two persons in Fig. \ref{fig0_2}(c) and (d) have similar ratios of the distance of the hip to the knee with that of the knee to the foot, since they are from the same identity.  Therefore, on the top of appearance features, extracting modality-invariant relation features may further enhance the discriminability of modality-shared features, and  reduce cross-modality variations and intra-modality variations for cross-modality person Re-ID. Here, we refer to the relations among different person parts as relation features.

Meanwhile, as shown in Fig. \ref{fig0_1}, we also find that one channel of appearance features extracted by using a Re-ID model from a person may mainly focus on describing one certain part of the person. For examples, the channel of appearance features in Fig. \ref{fig0_1}(b) mainly focuses on the head of a person, and the channel of appearance features in Fig. \ref{fig0_1}(c) mainly focuses on the body of a person. Motived by such observations, the relation features about a person can be obtained by capturing the relations among different channels of the appearance features of the person. To this end, a novel cross-modality person Re-ID model is presented in this paper for cross-modality person Re-ID, where the modality-shared appearance features and their relations among different channels are simultaneously adopted to boost performance. 

Concretely, a novel Multi-level Two-streamed Modality-shared Feature Extraction (MTMFE) sub-network is designed to simultaneously extract modality-shared appearance features and modality-invariant relation features. In MTMFE sub-network, modality-shared appearance features are first extracted by employing a shared 2D CNN based sub-network. Then, inspired by \cite{r84,r85}, modality-invariant relation features are extracted by using a shared 3D CNN, which has proven capability in capturing the relations of different frames in the 3D feature space \cite{r53,r54,r52}. In light of that,  in MTMFE sub-network, the extracted appearance features, including modality-specific ones and modality-shared ones, are considered as different `frames' and fed into the shared 3D CNN based sub-network to capture their relations. More specifically, the extracted modality-specific features and modality-shared appearance features are first projected into a shared 3D feature space, followed by a shared 3D CNN  based sub-network to capture the modality-invariant relation features. With the proposed MTMFE subnetwork, the discriminative of the extracted modality-shared features, including modality-shared appearance features and modality-invariant relation features, will be significantly enhanced for reducing the cross-modality variations and intra-modality variations, which will further boosts the performance of cross-modality person Re-ID. 

Besides, many works employ the Bi-directional Dual-constrained Top-Ranking (BDTR) loss or its variants \cite{r5, r17, r10, r31} to train their cross-modality person Re-ID models. The underlying constraint imposed by this sort of losses is that the distance of an anchor sample to its farthest cross-modality positive sample in the feature space should be smaller than the anchor sample to its nearest cross-modality negative sample by a predefined margin. However, such constraints cannot prevent the undesired situation that the distance of an anchor sample to its farthest cross-modality positive sample in the feature space is larger than the anchor sample to its nearest intra-modality negative sample. To address such issue, a novel Cross-modality Quadruplet (CQ) loss is designed, which simultaneously considers the following two aspects. Firstly, the distance of an anchor sample to its farthest cross-modality positive sample should be smaller than the anchor sample to its nearest cross-modality negative sample by a predefined margin. Secondly, the distance of an anchor sample to its farthest cross-modality positive sample should also be smaller than the anchor sample to its nearest intra-modality negative sample by a predefined margin. By virtue of the proposed CQ loss, more cross-modality constraints are introduced to train the proposed model, thus leading to better person Re-ID results.

In summary, the main contributions of this paper are as follows.

(1) As a departure from most existing models that rely only on the modality-shared appearance features, our new end-to-end cross-modality person Re-ID model makes use of both modality-shared appearance features and modality-invariant relation features to boost performance.  

(2) An MTMFE sub-network is designed to extract the modality-shared appearance features and the modality-invariant relation features in a shared 2D feature space and a shared 3D feature space, respectively. Doing so aims to reduce the cross-modality variations as well as the intra-modality variations. 

(3) A novel CQ loss is proposed to get the cross-modality variations further decreased by introducing more cross-modality constraints. On top of  the constraints among an anchor sample from one modality and its positive as well as negative samples from another modality, our proposed CQ loss further imposes  the constraints among an anchor sample from one modality, its positive samples from another modality and its negative sample from the same modality.

% \label{sec::rw} \label{sec::ex} \label{sec::con}

The rest of this paper is organized as follows. We briefly describe some previous works related to the visible image based person Re-ID and cross-modality person Re-ID first in Section \ref{sec::rw}, followed by the details of our newly proposed method in Section \ref{sec::pm}. Several experiments are conducted to validate the proposed model in Section \ref{sec::ex}, Finally, in Section \ref{sec::con}, a brief conclusion is made. 

\begin{figure*}[!t]
	\centering
	\includegraphics[width= \textwidth]{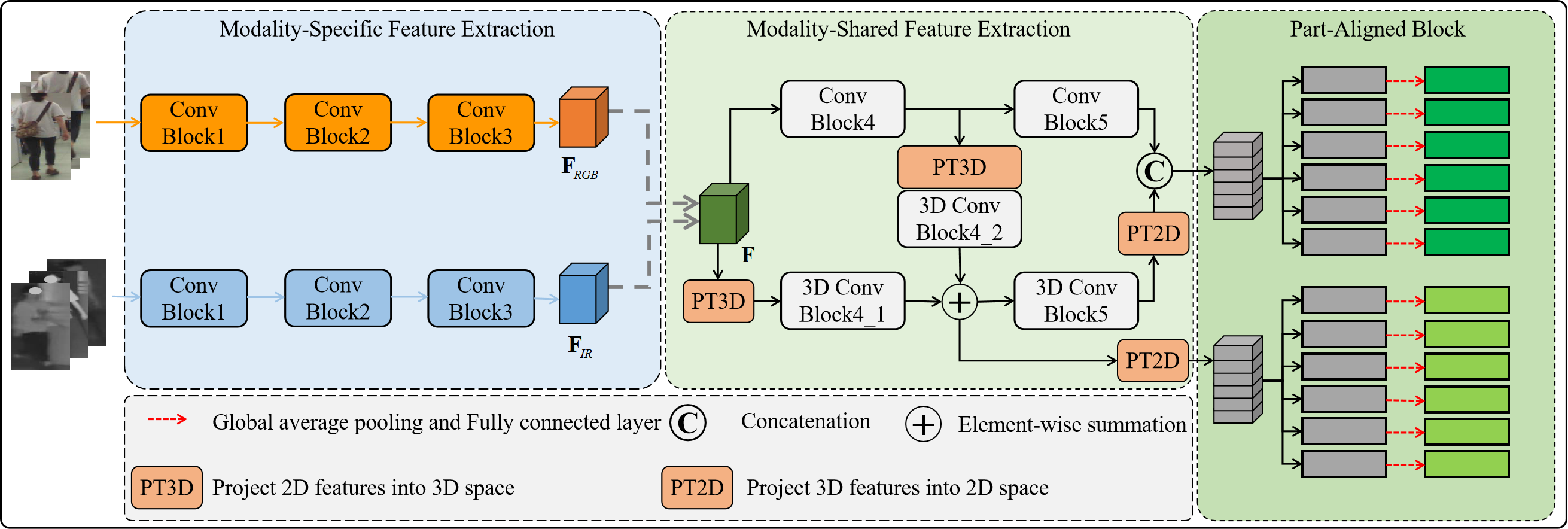}
	\caption{Illustration of the proposed model. A query RGB or IR image is first fed into the modality-specific feature extraction to extract corresponding modality-specific features. Then, the extracted modality-specific features are fed into the modality-shared feature extraction to extract modality-shared features, which contain the modality-shared appearance features and modality-invariant relation features. Finally, the extracted modality-shared features are fed into a part-aligned block to obtain the person features from different person parts. }
	\label{fig1}
\end{figure*}

\section{Related Work} \label{sec::rw} 

\subsection{Single-modality Person Re-ID}

Conventional person Re-ID models are mainly based on handcrafted features (\emph{e.g.,} color, textures, edges and shapes), followed by supervised distance metric learning \cite{r34, r35}. Recently, deep learning based person Re-ID models have become the mainstream and achieved great improvements over conventional ones. Generally, most existing deep learning based person Re-ID models can be divided into two categories: representation learning based models \cite{r36,r37,r38,r39,r46} and metric learning based models \cite{r41,r42,r43,r44,r45}. 

Representation learning based models \cite{r36,r37,r38,r39} try to obtain discriminative and robust person features from the input images. The extracted person features should be invariant to variations in illuminations, poses and viewpoints. For examples, a deep representation learning procedure, named as part loss network, was presented in \cite{r46} to learn discriminative representations for unseen person images by employing a novel part loss. Concretely, this part loss enforces their network to learn representations from different body parts and further gains the discriminative power on unseen persons. In \cite{r36}, a Pose-Invariant Embedding (PIE) was presented to align pedestrians to address the issue of pedestrian misalignment in person Re-ID. Specifically, it employs a pose estimator based PoseBox structure to produce well-aligned pedestrian images, so that the learned features can find the same person under intensive pose changes.  

Metric learning based models \cite{r41,r42,r43,r44,r45} determine whether two images are captured from the same pedestrian by measuring the similarity between their corresponding image features. For example, in \cite{r43}, a dedicated variant of the triplet loss was presented to perform end-to-end deep metric learning, providing guidance for triplet loss training. In \cite{r44}, a quadruplet loss was designed to improve the triplet loss, which would obtain features with a larger inter-class variations and a smaller intra-class variations.

\subsection{Cross-modality Person Re-ID}

Cross-modality person Re-ID aims to match the queries from one modality against a gallery set from another modality, which is very important for video surveillance in real-world scenarios. 

Previous methods can be summarized into two major categories: modality-shared feature learning and modality-specific feature compensation. Modality-shared feature learning based models  project modality-specific features into the same feature space \cite{r17,r16,r13,r12,r10,r9,r7,r10,r14,r8}, where cross-modality variations and intra-modality variations are simultaneously addressed by extracting discriminative modality-shared features. For example, in \cite{r17}, a dual-path network was presented, which first employs two AlexNet \cite{r75} based sub-networks to extract the modality-specific features from the input RGB and IR images, respectively. Then, they employ a shared sub-network with two stacked fully connected layers to transfer the extracted modality-specific features into a shared feature embedding for cross-modality person Re-ID. In \cite{r7}, besides a cross-modality triplet loss, a cutting-edge generative adversarial training based discriminator was also employed to learn more discriminative modality-shared feature representations from different modalities. Recently, apart from modality-shared features, modality-specific features are also employed to boost the performance of cross-modality person Re-ID. In \cite{r10}, a novel cross-modality shared-specific feature transfer algorithm was presented to explore the potential of both the modality-shared information and the modality-specific characteristics for boosting cross-modality person Re-ID.

Modality-specific feature compensation based models aim to generate the missing specific information from the existing ones \cite{r30, r8}, which address cross-modality variations by generating the images of missed modality and address cross-modality variations by extracting discriminative cross-modality features from paired RGB and IR images. For example, in \cite{r30}, a Alignment Generative Adversarial Network (AlignGAN) was designed, which consists of a pixel generator, a feature generator and a joint discriminator to jointly exploit pixel alignment and feature alignment. In \cite{r8}, a GAN based cross-modality person Re-ID model was presented  to generate the cross-modality paired-images by disentangling features and decoding from exchanged features. It also proposes a new variation module to map modality-invariant features to a latent manifold feature space for boosting cross-modality person Re-ID.

\subsection{3D Convolutional Neural Network}

3D Convolutional Neural Networks (3D CNNs) have been widely used in many computer vision tasks, \emph{e.g.,} video classification \cite{r49, r64}, action recognition \cite{r50, r52,r53, r85} and medical image processing \cite{r66, r65}. In video analysis tasks, 3D CNN can capture the relations among different video frames in the 3D feature space. In \cite{r53}, 3D CNN was first introduced to learn discriminative features along both spatial and temporal dimensions for action recognition. Then, in \cite{r52}, a 3D CNN, named as C3D, was designed to extract features for video processing, which achieves state-of-the-art performance on multiple video analysis tasks. In medical image processing, 3D CNN can capture the relations among different slices of the CT or MR images. For example, in \cite{r66}, a 3D CNN based Multiple Sclerosis (MS) lesion segmentation model was designed, which consists of two stages for automatically segmenting MS lesions. The alternative lesion voxels were selected in the first stage, while in the second stage, the final lesion voxels were segmented from the lesion voxels. 

In this paper, inspired by \cite{r84,r85}, 3D CNN is employed to extract the spatial and channel-wise relations among different appearance features. For that, the extracted appearance features are projected into a shared 3D feature space, where different appearance features are considered as different `video frames' or `CT/MR image slices' and fed into a shared 3D CNN based sub-network to extract the spatial and channel-wise relations among different appearance features. 

\section{Proposed Model} \label{sec::pm}

As shown in Fig. \ref{fig1}, the proposed model mainly contains three stages, \emph{i.e.,} modality-specific feature extraction, modality-shared feature extraction and part-aligned block. In modality-specific feature extraction, a two-streamed sub-network is first employed to extract modality-specific features from the input RGB image and IR images, respectively. Then, in modality-shared feature extraction, a Multi-level Two-streamed Modality-shared Feature Extraction (MTMFE) sub-network is presented to extract the modality-shared features, including modality-shared appearance features and modality-invariant relation features, for cross-modality person Re-ID. Finally, a part-aligned block is employed to extract final person features from different parts. More details about these three components will be discussed in the following contents.

\subsection{Modality-Specific Feature Extraction}

There are essential distinctions between RGB images and IR images due to the fact that RGB images and IR images are captured in different spectrums. As a result, the features extracted from RGB images and IR images have large variations. Considering that, two sub-networks with the same structures but different parameters are employed to extract modality-specific features from the given RGB images and IR images, respectively. One of the two sub-networks aims to extract modality-specific features from RGB images (denoted as $\mathbf{F}_{RGB}$) and the other is used to extract modality-specific features from IR images (denoted as $\mathbf{F}_{IR}$). Meanwhile, features at different levels have different properties. Low-level features contain more spatial or local-context information (\emph{e.g.,} colors, textures, edges, and contours), while high-level features contain more semantic information (\emph{e.g.,} objects and human parts). Furthermore, compared with high-level semantic information, low-level spatial information is more modality-related. For example, the semantic information of person bodies can be simultaneously contained in the high-level features of RGB images and IR images, while the color information only exists in the low-level features from RGB images. Therefore, in this paper, the two sub-networks employ a relatively shallower CNN for modality-specific feature extraction. Specifically, the two sub-networks follow the same structures as the first three convolutional blocks in ResNet50 \cite{r20} for modality-specific feature extraction, due to the fact that the first three convolutional blocks of ResNet50 mainly extract low-level features from the input images. In this way, the extracted modality-specific features can well exploit those single-modality information from the input images. Mathematically, this process is expressed by:
\begin{equation}\label{eq1}
\mathbf{F}_{RGB} = \operatorname{Conv}(X_{RGB},\theta_{RGB}),
\mathbf{F}_{IR} = \operatorname{Conv}(X_{IR},\theta_{IR}),
\end{equation} 
where $\operatorname{Conv}(*,\theta_{RGB})$ and $\operatorname{Conv}(*,\theta_{IR})$ denote the convolutional blocks with their corresponding parameters $\theta_{RGB}$ and $\theta_{IR}$, respectively. $X_{IR}$ and $X_{RGB}$ denote the input RGB image and IR image, respectively.

\subsection{Modality-shared Feature Extraction}

Modality-shared feature extraction aims to extract discriminative modality-shared features to reduce cross-modality variations and intra-modality variations. However, as discussed in Section \ref{sec::intro}, most existing models only extract modality-shared appearance features \cite{r17,r16} for cross-modality person Re-ID, which may not capture enough modality-invariant and discriminative information from RGB and IR images due to their large modality differences. Furthermore, the relations among different person parts (\emph{i.e.,} modality-invariant relation features) may be complementary to the modality-shared appearance features. Therefore, jointly capturing modality-shared appearance features and modality-invariant relation features may enhance the discriminability of the final modality-shared features. This further reduces cross-modality variations and intra-modality variations for cross-modality person Re-ID. To this end, a novel Multi-level Two-streamed Modality-shared Feature Extraction (MTMFE) sub-network is designed. 

As shown in Fig. \ref{fig1}, the proposed MTMFE sub-network also contains two shared sub-networks. One is a 2D CNN based sub-network to extract the modality-shared appearance features by projecting the extracted modality-specific features into a shared 2D feature space. The other is a 3D CNN based sub-network to capture the modality-invariant relation features by projecting the extracted modality-specific features and the modality-shared appearance features into a shared 3D feature space. It should be noted that the modality-specific features for modality-shared feature extraction may be extracted from the RGB image (\emph{i.e.,} $\mathbf{F}_{RGB}$) or the IR image (\emph{i.e.,} $\mathbf{F}_{IR}$). For simplicity, the modality-specific features (\emph{i.e.,} $\mathbf{F}_{RGB}$ or $\mathbf{F}_{IR}$) for modality-shared feature extraction  are denoted by $\mathbf{F} \in R^{C\times H \times W}$ in this section. Here, $C, H $ and $W$ denote its channels, height and width.

\subsubsection{Modality-shared Appearance Feature Extraction}

As shown in Fig. \ref{fig1}, on the top of modality-specific feature extraction, a shared 2D CNN based sub-network is first employed to extract  modality-shared appearance features by projecting the modality-specific features into a shared feature space. Specifically, given the modality-specific features $\mathbf{F}$, the shared sub-network employs two convolutional blocks to extract two levels of modality-shared appearance features (denoted as $\mathbf{F}_{a}^1$ and $\mathbf{F}_{a}^2$, respectively). 
Mathematically, this process is expressed by:
\begin{equation}\label{eq2}
\mathbf{F}_{a}^1 = \operatorname{Conv}(\mathbf{F},\theta_{a1}),
\mathbf{F}_{a}^2  = \operatorname{Conv}(\mathbf{F}_{a}^1 ,\theta_{a2}),
\end{equation} 
where $\operatorname{Conv}(*,\theta_{a1})$ and $\operatorname{Conv}(*,\theta_{a2})$ denote two convolutional blocks with their corresponding parameters $\theta_{a1}$ and $\theta_{a2}$, respectively. In our model, the shared sub-network follows the structure of the last two convolutional blocks of ResNet50 \cite{r20}. In our model, both the two levels of modality-shared appearance features ($\mathbf{F}_{a}^1$ and $\mathbf{F}_{a}^2$) are employed for cross-modality person Re-ID, due to the fact that the modality-shared appearance features in the two levels contain varieties of semantic information. This may further enhance the discriminability and robustness of the final person features. 

\subsubsection{Modality-shared relation Feature Extraction}

\begin{figure}[!t]
	\centering
	\includegraphics[width= 0.5\textwidth]{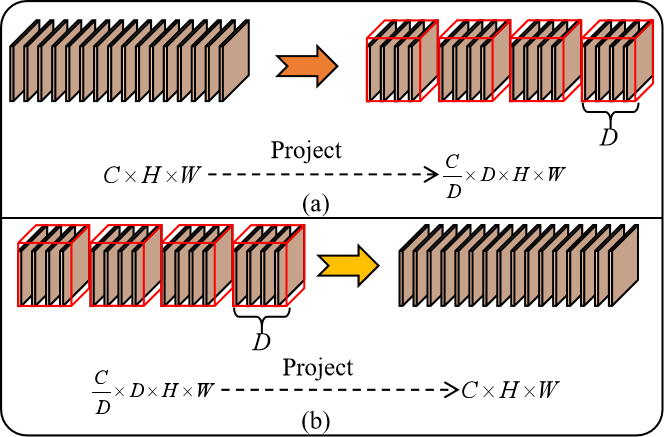}
	\caption{Illustration of the projection operation. (a) Project the 2D appearance features into 3D feature space; (b) Project the modality-invariant relation features into 2D feature space. In (a), the input appearance features with the size of $C\times H \times W$ are projected as 3D features with the size of $\frac{C}{D} \times D \times H \times W$. In (b), the input relation feature with the size of  $\frac{C}{D} \times D \times H \times W$ are projected as 2D features with the size of $C\times H \times W$. Here, $C, H, W$ and $D$ denote channels, height, width and depth of corresponding features.}
	\label{fig4}
\end{figure}

As discussed in Section \ref{sec::intro}, extracting modality-invariant relation features will further enhance the discriminability of modality-shared features and reduce cross-modality variations as well as intra-modality variations for cross-modality person Re-ID. Furthermore, different channels of appearance features mainly contain discriminative information at different person parts. Therefore, relation features can be obtained by capturing the relations among different channels of appearance features. Considering that, in our proposed model, the next step is to capture modality-invariant relation features from different types of appearance features (\emph{i.e.,} modality-specific features and modality-shared appearance features).  

%As discussed in Section \ref{sec::intro}, extracting modality-invariant relation features may further enhance the discriminability of modality-shared features and reduce cross-modality variations as well as intra-modality variations for cross-modality person Re-ID. Furthermore, appearance features mainly contain discriminative information at different person parts. Therefore, relation features can be obtained by capturing the relations among different channels of appearance features. Considering that, in our proposed model, we capture the modality-invariant relation features from different types of appearance features (\emph{i.e.,} modality-specific features and modality-shared appearance features). 

%As discussed in Section \ref{sec::intro}, modality-invariant relation features may also contain much modality-invariant and discriminative information, which are complement to those modality-shared appearance features. Furthermore,  Therefore, on the top of modality-shared appearance features, extracting the relation features may further enhance the discriminability of modality-shared features and reduce cross-modality variations as well as intra-modality variations. Considering that, in our proposed model, the next step is to capture the modality-invariant relation features. 

As shown in Fig. \ref{fig1}, in the proposed MTMFE sub-network, two levels of the relation features (\emph{i.e.,} $\mathbf{F}_{I}^{1}$ and $\mathbf{F}_{I}^{2}$) are extracted. Moreover, the first-level modality-invariant relation features $\mathbf{F}_{I}^1$  are composed of two sub-parts (\emph{i.e.,} $\mathbf{F}_{I}^{1\_1}$ and $\mathbf{F}_{I}^{1\_2}$). Here, $\mathbf{F}_{I}^{1\_1}$ is extracted from the modality-specific features $\mathbf{F}$, while $\mathbf{F}_{I}^{1\_2}$ is extracted from the first level of modality-shared appearance features $\mathbf{F}_{a}^1$. The second-level modality-invariant relation features $\mathbf{F}_{I}^{2}$ are directly extracted from the first level of modality-invariant relation features $\mathbf{F}_{I}^{1}$. 

\begin{figure}[!t]
	\centering
	\includegraphics[width= 0.5\textwidth]{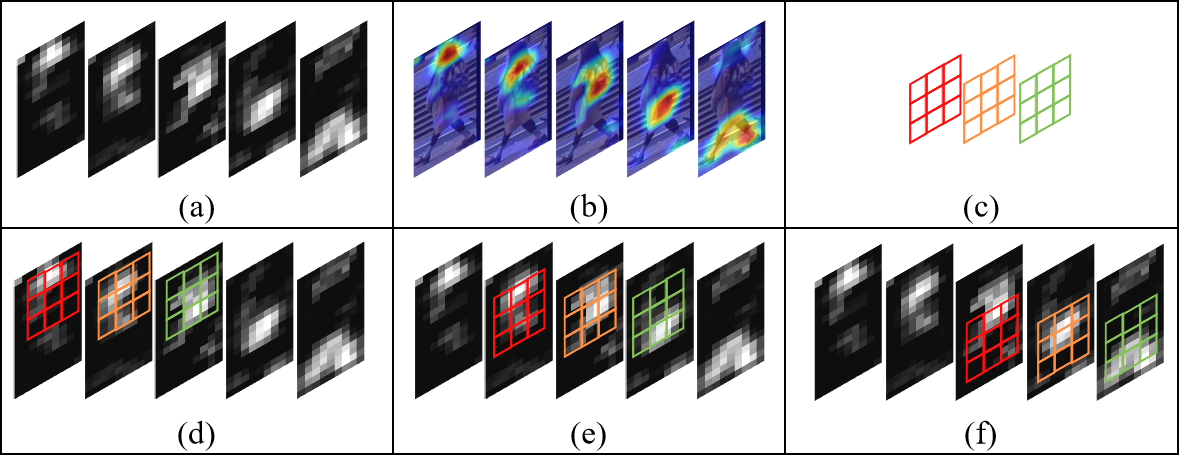}
	\caption{Illustration of the extraction of the modality-invariant relation features. (a) A 3D features transferred from the modality-shared appearance features. (b) Corresponding heat maps of the modality-shared appearance features. (c) A 3D convolutional kernel. (d)-(f) Capturing the modality-invariant relation features at different spatial locations by using the 3D convolutional kernel. It can be seen that the spatial-wise and channel-wise relation of the different appearance features are simultaneously captured by using the 3D convolutional kernel. As a result, the relation of those spatial information of different person parts are effectively extracted to reduce the cross-modality variants and further enhance the discriminability of the modality-shared features for cross-modality person Re-ID. }
	\label{fig13}
\end{figure}

Specifically, for the extraction of $\mathbf{F}_{I}^{1\_1}$, inspired by \cite{r84,r85}, the modality-specific features $\mathbf{F}$ are first projected into a shared 3D feature spaces by using the proposed Projecting features inTo 3D feature space (PT3D) operation. Then, a 3D convolutional block is employed to extract the relations among the modality-specific features. Here, as shown in Fig. \ref{fig4}(a), the PT3D operation is implemented by first splitting the extracted modality-specific features $\mathbf{F}$ into multiple groups along its channel dimension, where each group contains $D$ channels of the features (\emph{i.e.,} total $\frac{C}{D}$ groups). Then, each group is constructed as a 3D feature of size  $D \times H \times W$. In this way, the extracted modality-specific features are projected into a shared 3D feature space. Here, as in \cite{r84,r85}, the 3D feature space is actually a pseudo 3D feature space due to the fact that the features are not extracted from the real 3D data, liking videos or medical images, but projected from the extracted 2D features. 

In this shared 3D feature space, the spatial- and channel-wise relations among the modality-specific features are simultaneously established for cross-modality person Re-ID by using two stacked 3D convolutional layers with kernel size of $3 \times 3 \times 3$.  For better understanding, Fig. \ref{fig13}(a) can be seen as a 3D feature transferred from a group of modality-shared appearance features. Fig. \ref{fig13}(b) demonstrates that the modality-shared appearance features contain the spatial information of different person parts. As shown in Fig. \ref{fig13}(d)-(f), the relations of the modality-shared appearance features corresponding to different person parts can be captured by performing a 3$\times$3$\times$3 3D convolutional kernel (\emph{e.g.,} Fig. \ref{fig13}(c)) in the shared 3D feature space. Furthermore, more modality-invariant relation features can be extracted by employing more 3D convolutional kernels. Mathematically, this process is expressed by:
\begin{equation}\label{eq7}
\mathbf{F}_{I}^{1\_1} = \operatorname{Conv3d}(\operatorname{PT3D}(\mathbf{F}), \delta_1),
\end{equation} 
where $\operatorname{Conv3d}(*, \delta_1)$ denotes two stacked 3D convolutional layers with their corresponding parameters $\delta_1$. $\operatorname{PT3D}(*)$ denotes the projection operation.

Then, considering that the extracted modality-specific features and modality-shared appearance features have different properties, the relations of the modality-shared appearance features are also captured to increase the diversity of the modality-invariant relation features and further enhance the discriminability of modality-shared features for cross-modality person Re-ID in this paper. For that, the second part of the  modality-invariant relation features in the first level (\emph{i.e.,} $\mathbf{F}_{I}^{1\_2}$) are extracted from the first level of modality-shared appearance features (\emph{i.e.,} $\mathbf{F}_{a}^1$) in the same way as in the previous step. Concretely, $\mathbf{F}_{a}^1$ is first projected into the shared 3D feature space by using the proposed PT3D operation and then fed into another two stacked 3D convolutional layers to extract $\mathbf{F}_{I}^{1\_2}$. Mathematically, this process is expressed by:
\begin{equation}\label{eq7_1}
\mathbf{F}_{I}^{1\_2} = \operatorname{Conv3d}(\operatorname{PT3D}(\mathbf{F}_{a}^1), \delta_2),
\end{equation} 
where $\operatorname{Conv3d}(*, \delta_2)$ denotes the two stacked 3D convolutional layers with its corresponding parameters $\delta_2$. 

Finally, the first level of modality-invariant relation features $\mathbf{F}_{I}^{1}$ are obtained by the element-wise summation of $\mathbf{F}_{I}^{1\_1}$ and $\mathbf{F}_{I}^{1\_2}$, \emph{i.e.,}
\begin{equation}\label{eq8}
\mathbf{F}_{I}^{1} = \mathbf{F}_{I}^{1\_1} + \mathbf{F}_{I}^{1\_2}.
\end{equation} 
Here, the element-wise summation is employed for its relatively lower computational complexity and memory usage. Other fusion strategies (\emph{e.g.,} concatenation) may also be used. 

The second level of modality-invariant relation features $\mathbf{F}_{I}^{2}$ are obtained by feeding $\mathbf{F}_{I}^{1}$ into the second 3D convolutional block, which is expressed by
\begin{equation}\label{eq9}
\mathbf{F}_{I}^{2} = \operatorname{Conv3d}(\mathbf{F}_{I}^{1}, \delta_3),
\end{equation} 
where  $\operatorname{Conv3d}(*, \delta_3)$ denotes the 3D convolutional block with its corresponding parameters $\delta_3$. Furthermore, the relations among the second-level modality-shared appearance features $\mathbf{F}_{a}^2$ may also be extracted by projecting the second-level modality-shared appearance features into the shared 3D feature space. However, our experimental results show that the performance of cross-modality person Re-ID may not be further improved even if more modality-invariant relation features are extracted. Therefore, the modality-invariant relation features are not further extracted from the second-level of appearance features $\mathbf{F}_{a}^2$ in our proposed model.

In this way, the relations of the modality-specific features and modality-shared appearance features at different person parts are captured in a shared 3D features space, respectively. By virtue of modality-shared appearance features and modality-invariant relation features, the diversity and discriminability of the final modality-shared features are significantly increased, thus effectively reducing cross-modality variations and intra-modality variations for cross-modality person Re-ID.  

Finally, the extracted modality-invariant relation features are projected back into the 2D feature space and concatenated with the modality-shared appearance features in the same levels to facilitate the subsequent steps and meet the requirement of the loss function. In this way, the modality-shared features in different levels are thus obtained, \emph{i.e.,}
\begin{equation}\label{eq9_1}
\mathbf{F}_{s}^{1} = \operatorname{Cat}(\mathbf{F}_{a}^1, \operatorname{PT2D}(\mathbf{F}_{I}^1)), \mathbf{F}_{s}^{2} = \operatorname{Cat}(\mathbf{F}_{a}^2, \operatorname{PT2D}(\mathbf{F}_{I}^2)),
\end{equation} 
where $\mathbf{F}_{s}^{1}$ and $\mathbf{F}_{s}^{2}$ denote the final modality-shared features. $\operatorname{PT2D}(*)$, as shown in Fig. \ref{fig4}(b), denotes the projection operation that projects the 3D features back into the 2D feature space, which is the reverse operation of $\operatorname{PT3D}(*)$. 

\subsection{Part-Aligned Block}

Given the extracted modality-shared features $\mathbf{F}_{s}^{1}$ and $\mathbf{F}_{s}^{2}$, a part-aligned block is employed to extract the final person features from different person parts. Specifically, as shown in Fig. \ref{fig1}, following the PCB-based model \cite{r61, r4}, the extracted modality-shared features $\mathbf{F}_{s}^{1}$ and $\mathbf{F}_{s}^{2}$ are separated into six parts, respectively. For the features within each part, a Global Average Pooling (GAP) is employed to obtain the global information of each part, Then, a fully connected layer is further employed to obtain the final feature embedding for representing the corresponding part of a given image. Mathematically, this process is expressed by:
\begin{equation}\label{eq10}
\mathbf{\hat{F}}_{p,1}^i,...,\mathbf{\hat{F}}_{p,6}^i = \operatorname{Part}(\mathbf{F}_{s}^{i}),
\end{equation}   
\begin{equation}\label{eq11}
\mathbf{F}_{p,k}^i = \operatorname{FC}(\operatorname{GAP}(\mathbf{\hat{F}}_{p,k}^i), \vartheta_k),
\end{equation}   
where $k$=1,2,3,4,5,6 denotes different parts and $i$=1,2 denotes different levels.  $\operatorname{Part}(*)$ denotes the separation operation. $\operatorname{FC}(*, \vartheta_k)$ denotes the fully connected layers with corresponding parameters $\vartheta_k$.  $\operatorname{GAP}(*)$ denotes the global average pooling. $\mathbf{F}_{p,k}^i$ and $\mathbf{\hat{F}}_{p,k}^i$ denote the features from $k$-th part of the $i$-th level features, respectively. 

Furthermore, the features $\mathbf{F}_{p,k}^i$ from different person parts should be discriminative for different persons. For that, the features $\mathbf{F}_{p,k}^i$ are employed to predict the corresponding person identification. Concretely, for the features within each part, an FC layer is performed to predict which person identification it belongs to. Mathematically, this process is expressed by:
\begin{equation}\label{eq11_1}
cls_{p,k}^i = \operatorname{FC}(\mathbf{F}_{p,k}^i, \vartheta_{c_k^i}),
\end{equation}   
where $cls_{p,k}^i $ denotes the person identification classification score derived from the $k$-th part in the $i$-th level of features. $\operatorname{FC}(*, \vartheta_{c_k^i})$ denotes the fully connected layers with the corresponding parameters $\vartheta_{c_k^i}$.

\subsection{Loss Function}
\begin{figure}[!t]
	\centering
	\includegraphics[width= 0.5\textwidth]{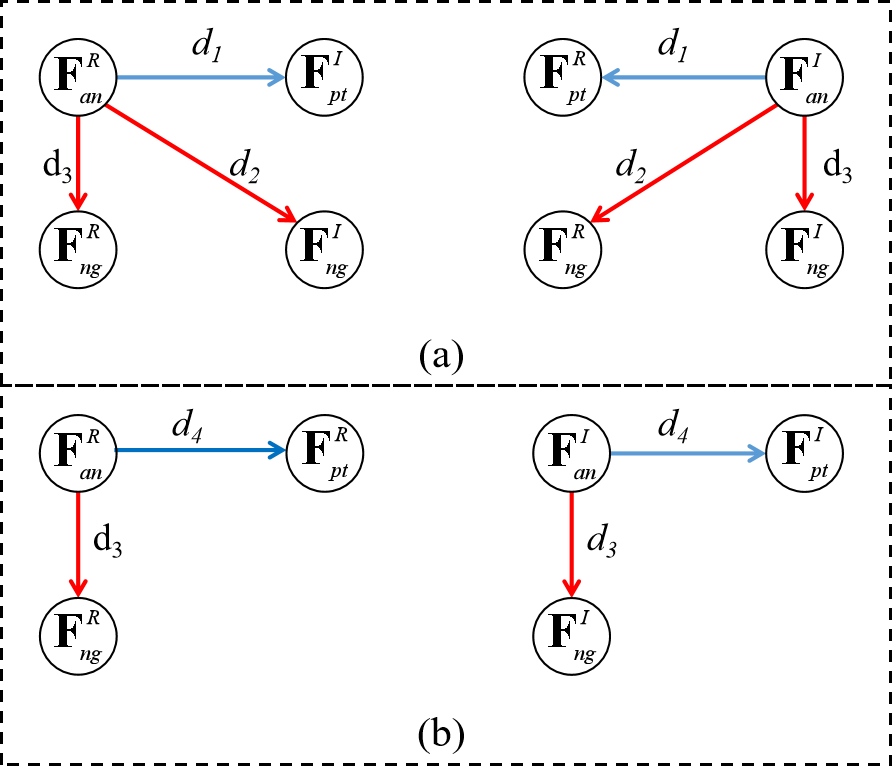}
	\caption{Illustration of the proposed cross-modality quadruplet loss. (a) The proposed cross-modality quadruplet loss; (b) Single-modality triplet loss. Here, $d_i, i=1,2,3,4$ denotes the distance of two features. For example, $d_1$ represents the distance of the features extracted from the images of anchor-positive pairs in different modalities. Meanwhile, the smaller value of $d_i$ represents that the two features become more similar.}
	\label{fig2}
\end{figure}

After extracting person features from the input visible or thermal images by employing the proposed network, a cross-modality quadruplet loss is further designed to supervise the feature learning objectives. Furthermore, the intra-modality variations are also reduced by employing a single-modality triplet loss in this paper. 

Suppose that, given an anchor sample, $\mathbf{F}_{an}^R$ and $\mathbf{F}_{an}^I$ denote its features extracted from the corresponding RGB and thermal images, respectively. Similarly, $\mathbf{F}_{pt}^R$ and $\mathbf{F}_{pt}^I$ denote the features extracted from corresponding positive RGB and thermal image samples, while $\mathbf{F}_{ng}^R$ and $\mathbf{F}_{ng}^I$ denote the features extracted from corresponding negative RGB and thermal image samples. The details of our losses are discussed as follows. 

\subsubsection{Cross-modality Quadruplet Loss}

Many existing works employ the Bi-directional Dual-constrained Top-Ranking (BDTR) loss or its variants \cite{r5, r17, r10, r31} to train their cross-modality person Re-ID models. Following the BDTR loss (\emph{i.e.,} shown in Eq. \ref{eq3}), these losses only constrain that the distance of an anchor sample to its farthest cross-modality positive sample should be smaller than the anchor sample to its nearest cross-modality negative sample by a predefined margin (\emph{i.e.,} $d_1<d_2$ in Fig. \ref{fig2}(a)).
\begin{equation}\label{eq3}
\begin{split}
{\zeta }_{bi}=&\sum \operatorname{max}[\rho_1+\operatorname{D}(\mathbf{F}_{an}^R, \mathbf{F}_{pt}^I)-\operatorname{D}(\mathbf{F}_{an}^R, \mathbf{F}_{ng}^I)] \\ &+ \sum\operatorname{max}[\rho_1+\operatorname{D}(\mathbf{F}_{an}^I, \mathbf{F}_{pt}^R)-\operatorname{D}(\mathbf{F}_{an}^I, \mathbf{F}_{ng}^R)],
\end{split}
\end{equation}
where  $\rho_1$ denotes the corresponding margin. $\operatorname{D}(*)$ denotes the distance function, which default is the squared Euclidean distance, \emph{i.e.,} 
\begin{equation}\label{eq5}
\operatorname{D}(\mathbf{X},\mathbf{Y}) = \frac{1}{2}\parallel \mathbf{X}-\mathbf{Y} \parallel^2_2,
\end{equation}
where $\mathbf{X}$ and $\mathbf{Y}$ denote the input features.

However, as shown in Fig. \ref{fig2}(a), the cross-modality discrepancy may not be fully reduced by only employing the BDTR loss or its variants. They may ignore another aspect that is also caused by modality differences, \emph{i.e.,} the distance of an anchor sample to its farthest cross-modality positive sample may also be larger than the anchor sample to its nearest intra-modality negative sample (\emph{i.e.,} $d_1>d_3$ in \ref{fig2}(a)). Considering that, a novel Cross-modality Quadruplet (CQ) loss is designed to further reduce the cross-modality discrepancy by simultaneously considering the following two aspects. (1) The distance of an anchor sample to its farthest cross-modality positive sample should be smaller than the anchor sample to its nearest cross-modality negative sample by a predefined margin (\emph{i.e.,} $d_1<d_2$). (2) The distance of an anchor sample to its farthest cross-modality positive sample should also be smaller than the anchor sample to its nearest intra-modality negative sample by a predefined margin (\emph{i.e.,} $d_1<d_3$).  This is expressed by:
\begin{equation}\label{eq4}
\begin{split}
{\zeta }_{cq}=& \sum\operatorname{max}[\rho_2+\operatorname{D}(\mathbf{F}_{an}^R, \mathbf{F}_{pt}^I)-\operatorname{D}(\mathbf{F}_{an}^R, \mathbf{F}_{ng}^I)] \\ &+  \sum\operatorname{max}[\rho_2+\operatorname{D}(\mathbf{F}_{an}^I, \mathbf{F}_{pt}^R)-\operatorname{D}(\mathbf{F}_{an}^I, \mathbf{F}_{ng}^R)] \\ & +  \sum\operatorname{max}[\rho_2+\operatorname{D}(\mathbf{F}_{an}^R, \mathbf{F}_{pt}^I)-\operatorname{D}(\mathbf{F}_{an}^R, \mathbf{F}_{ng}^R)] \\ & +  \sum\operatorname{max}[\rho_2+\operatorname{D}(\mathbf{F}_{an}^I, \mathbf{F}_{pt}^R)-\operatorname{D}(\mathbf{F}_{an}^I, \mathbf{F}_{ng}^I)].
\end{split}
\end{equation}
Here, $\rho_2$ denotes the corresponding margin. Note that all the input features are $l_2$ normalized for stable convergence. 

\subsubsection{Single-modality Triplet Loss}

Besides the proposed CQ loss, single-modality triplet loss \cite{r43,r44} is also employed to reduce the large intra-modality variations. As shown in Fig. \ref{fig2}(b), single-modality triple loss tries to let the distance of an anchor sample to its farthest intra-modality positive sample smaller than the anchor sample to its nearest intra-modality negative sample by a predefined margin. This is expressed by
\begin{equation}\label{eq6}
\begin{split}
{\zeta }_{st}=& \sum\operatorname{max}[\rho_3+\operatorname{D}(\mathbf{F}_{an}^R, \mathbf{F}_{pt}^R)-\operatorname{D}(\mathbf{F}_{an}^R, \mathbf{F}_{ng}^R)] \\ &+  \sum\operatorname{max}[\rho_3+\operatorname{D}(\mathbf{F}_{an}^I, \mathbf{F}_{pt}^I)-\operatorname{D}(\mathbf{F}_{an}^I, \mathbf{F}_{ng}^I)].
\end{split}
\end{equation}    
Here, $\rho_3$ denotes the corresponding margin.

\subsubsection{Identification loss}

Finally, an identification loss is employed to make sure that those extracted features are discriminative for pedestrians, which is expressed by:
\begin{equation}\label{eq12}
{\zeta }_{id} = -\frac{1}{N_{id}}\sum_{c=1}^{N_{id}}q_c\operatorname{log}(p_c),
\end{equation}  
where $N_{id}$ denotes the total numbers of person identifications.  $q_c$ and $p_c$ denote the ground truth and the predicted probability of being the $c$-th identification, respectively. In our proposed model, for features of each part in each level, the identification loss is employed to facilitate the proposed model to learn more discriminative person features.

\textbf{Total Loss for Training:} In the training process, the anchor samples ($I_{an}^{R}$ and $I_{an}^{I}$), positive samples ($I_{pt}^{R}$ and $I_{pt}^{I}$) and negative samples ($I_{ng}^{R}$ and $I_{ng}^{I}$) are sampled. After feeding these samples into the proposed model, corresponding person part features are obtained, \emph{i.e.,} $\mathbf{F}_{an,k}^{R,i}$, $\mathbf{F}_{an,k}^{I,i}$, $\mathbf{F}_{pt,k}^{R,i}$, $\mathbf{F}_{pt,k}^{I,i}$, $\mathbf{F}_{ng,k}^{R,i}$ and $\mathbf{F}_{ng,k}^{I,i}$. Here, $k$=1,2,3,4,5,6 denotes different parts and $i$=1,2 denotes different levels. Meanwhile, the classification scores for the anchor images (\emph{i.e.,} $p_{an,k}^{R,i}$ and $p_{an,k}^{I,i}$) are also obtained. Based on that, the total loss ${\zeta }$ for the anchor images in the training process is expressed by
\begin{equation}\label{eq13}
\begin{split}
{\zeta } = \sum^{2}_{i=1}&\sum^{6}_{k=1}({\zeta }_{cq}(\mathbf{F}_{an,k}^{R,i}, \mathbf{F}_{an,k}^{I,i}, \mathbf{F}_{pt,k}^{R,i}, \mathbf{F}_{pt,k}^{I,i}, \mathbf{F}_{ng,k}^{R,i}, \mathbf{F}_{ng,k}^{I,i}) \\ & + {\zeta }_{st}(\mathbf{F}_{an,k}^{R,i}, \mathbf{F}_{an,k}^{I,i}, \mathbf{F}_{pt,k}^{R,i}, \mathbf{F}_{pt,k}^{I,i}, \mathbf{F}_{ng,k}^{R,i}, \mathbf{F}_{ng,k}^{I,i}) \\ &  + {\zeta }_{id}(p_{an,k}^{R,i}, q_{id})+ {\zeta }_{id}(p_{an,k}^{I,i}, q_{id})),
\end{split}
\end{equation}  
where $q_{id}$ denotes the ground truth person identification. 

\section{Experiments} \label{sec::ex}   

\subsection{Datasets and Evaluation Metrics}

\textbf{Datasets:} To evaluate the performance of our proposed network, we conduct evaluations on two public cross-modality person Re-ID benchmarks: SYSU-MM01 \cite{r15} and RegDB \cite{r57}.

SYSU-MM01 is a large-scale and widely used cross-modality person Re-ID dataset \cite{r15}. Four visible cameras and two infrared cameras are employed to collect RGB images and IR images from both indoor and outdoor environments. Its training set contains 395 person identities, including totally 22,258 RGB images and 11,909 IR images. Furthermore, the testing set of SYSU-MM01 contains 96 person identities with 3,803 IR images for query and 301/3010 (one-shot/multi-shot) randomly selected RGB images as the gallery. There are two evaluation modes in SYSU-MM01 for cross-modality person Re-ID: indoor-search and all-search \cite{r15}.

RegDB is another widely used cross-modality person Re-ID dataset \cite{r57}. It totally contains 412 person identities and 8240 images collected by dual camera systems, where 206 identities are employed for training and the rest of 206 identities are employed for testing. For each person identity, 10 RGB images and 10 IR images under different poses, positions and illuminations are available. There are also two evaluation modes for RegDB. One is RGB-IR matching for searching IR images from a given RGB images. The other mode is IR-RGB matching for searching RGB images from a given IR image.

\textbf{Evaluation metrics:} Following existing works \cite{r4, r5, r7, r67}, Cumulated Matching Characteristics (CMC) and mean Average Precision (mAP) are adopted as the evaluation metrics. Here, CMC (\emph{i.e.,} Rank-r accuracy) measures the probability of a correct cross-modality person image occurs in the top-r retrieved results. mAP measures the retrieval performance when multiple matching images occur in the gallery set. 

%the Rank-1, Rank-10 and Rank-20 accuracy as well as mean average precision (mAP) are adopted to validate our proposed model, since one person has multiple groundtruths in the gallery set.

\subsection{Online Batch Sampling Strategy}

Our online batch sampling strategy is discussed in this subsection. Specifically, for each batch in a training iteration, $N$ person identities are first selected from training sets randomly. Then, for each selected person identity, $K$ corresponding RGB images and $K$ IR images are randomly selected. As a result, each training batch contains $N\times K$ RGB images and $N\times K$ IR images (\emph{i.e.,} totally $2\times N\times K$ images) for training. Here, we set $N=8$ and $K=4$ during our training process.

In the testing strategy, the features of all the query images are first obtained by feeding all the query images into our model. Similarly, the features of all the gallery images are then obtained by feeding all the gallery images into our model. After that, the pairwise distances between the query and gallery images are calculated and the corresponding quantitative results are then obtained.

\subsection{Implementation details}

The proposed model is implemented on an NVIDIA 1080Ti GPU by using the PyTorch repository \cite{r59}. The parameters of the modality-specific extraction sub-networks and the 2D CNN of the modality-shared feature extraction sub-networks are initialized by using a pre-trained ResNet50 \cite{r20} on ImageNet \cite{r22}. Other parameters are randomly initialized by using the Xavier initialization \cite{r60}. All the experiments are performed by using the SGD optimizer with the weight decay of 0.0005. The initial learning rates for the parameters of modality-specific features extraction sub-networks are set to 0.01 and for those of modality-shared feature extraction sub-networks are set to 0.1. The learning rates are decreased by a factor of 0.1 for every 7 epochs. In the training process, all the sampled images are resized to the $288\times 144$ by employing the $\operatorname{bilinear}$ operation. Meanwhile, some data augmentations, such as random flipping and cropping, are also employed. 

\begin{figure}[!t]
	\centering
	\includegraphics[width= 0.5 \textwidth]{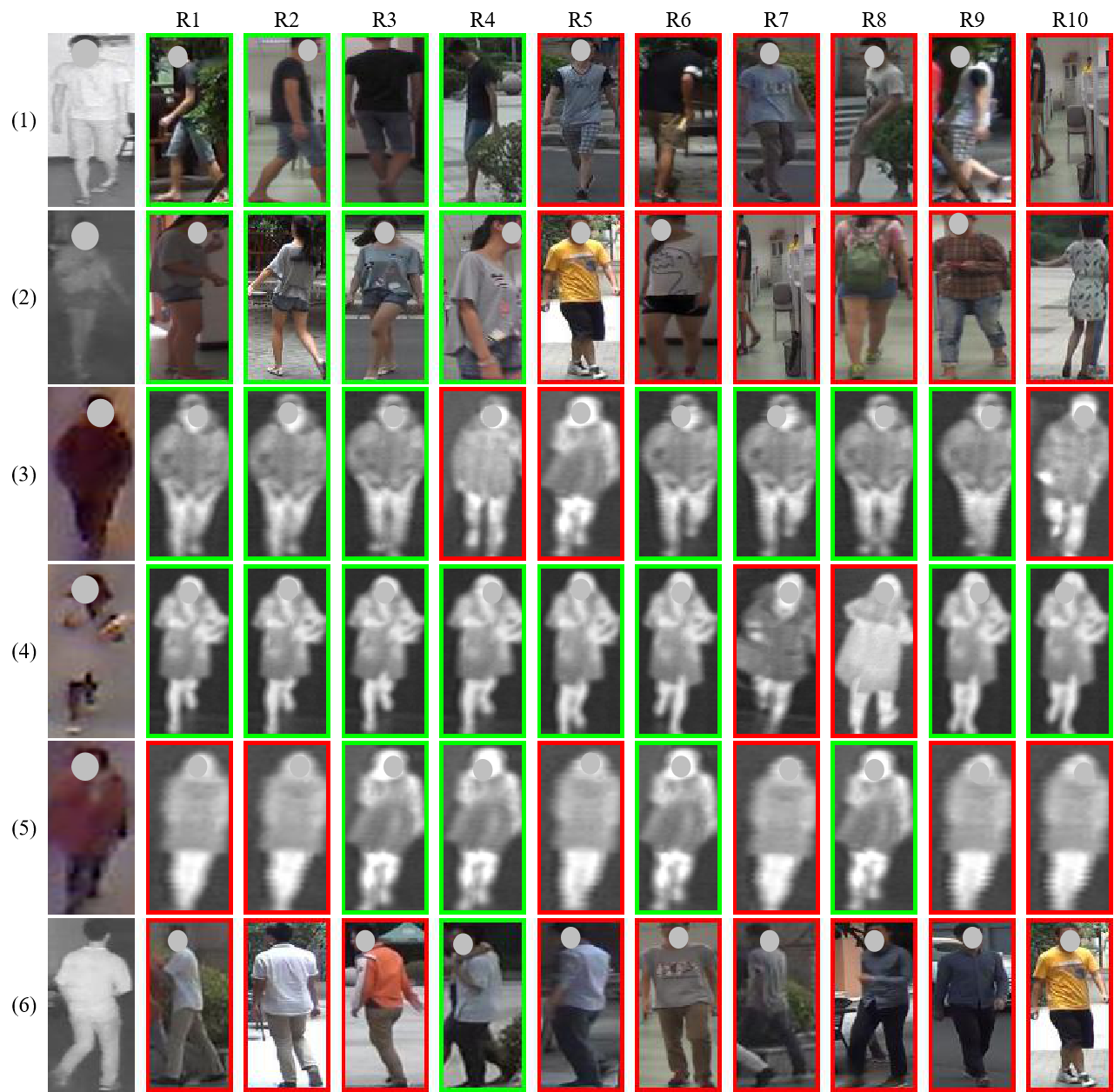}
	\caption{The top-10 retrieved results of some example queries with the proposed method on the SYSU-MM01 and RegDB datasets. The green bounding boxes indicate the correct matchings and red bounding boxes represent wrong matchings (best viewed in color). Examples in the first two rows are sampled from SYSU-MM01 dataset which takes the IR images as queries to search from the RGB images. Examples in the third and fourth rows are sampled from RegDB dataset which takes RGB images as queries to search from IR images. The last two rows show some failure cases.}
	\label{fig3}
\end{figure}

\subsection{Visualization}

The top ten retrieved results of some randomly selected query examples on the SYSU-MM01 and RegDB datasets are shown in Fig. \ref{fig3}. We can see that cross-modality person Re-ID is an extremely challenging task and it is even hard for human to tell which person is the correct matching of the query. Then, there are some wrongly retrieved examples in the ranking list, but the top-ranked images are usually quite close to the query image due to similar appearances and relation information. Meanwhile, as shown in the last two rows of Fig. \ref{fig3}, for those failure cases, the top matches in the ranking list are indeed very similar to the query person.

\subsection{Ablation study}

\begin{table}[!t]
	\renewcommand{\arraystretch}{1.3}
	\caption{quantitative results of different ablation experiments.}
	\label{tab1}
	\centering
	\begin{tabular}{ccccccccc}
		\hline	
		Methods	        & ML      & P      & RF-only     &RF    & CQ   & r1   & r10   &MAP   \\ 
		\hline
		  1    &$\times$  & $\times$ &  $\times$   & $\times$ & $\times$          & 16.43  & 51.84   & 18.50  \\
		  2    &\checkmark  & $\times$ &  $\times$   & $\times$ & $\times$        & 18.25  & 57.38   & 19.81  \\
		  3	   &$\times$  & \checkmark &  $\times$   & $\times$ & $\times$        & 45.36  & 85.22   & 45.28 \\	
		  4	   &\checkmark  & \checkmark&  $\times$   & $\times$ & $\times$       & 48.15  & 77.78   & 48.35  \\
		  5    &\checkmark  & \checkmark& \checkmark   & $\times$ & $\times$      & 47.17  & 88.25   & 46.29  \\
		  6    &\checkmark  & \checkmark& $\times$   & \checkmark & $\times$    &  60.66  & 91.61   & 58.28  \\
		  7    &\checkmark  & \checkmark& $\times$  & \checkmark & \checkmark   &\textbf{62.56} &\textbf{93.85}  &\textbf{60.57}  \\
		\hline
	\end{tabular}
\end{table}

\begin{table*}[!t]
	\renewcommand{\arraystretch}{1.3}
	\caption{Comparison with the state-of-the-art models on SYSU-MM01 dataset.}
	\label{tab2}
	\centering
	\resizebox{\textwidth}{!}{
		\begin{tabular}{|c|cccc|cccc|cccc|cccc|}
			\hline	
			-	  &  \multicolumn{8}{c|}{All-Search}  & \multicolumn{8}{c|}{Indoor-Search}  \\ 
			\hline	
			-	  &  \multicolumn{4}{c|}{Single-shot}  & \multicolumn{4}{c|}{Multi-shot} &  \multicolumn{4}{c|}{Single-shot}  & \multicolumn{4}{c|}{Multi-shot} \\
			\hline	
			Methods	  & R1   & R10   & R20   & mAP & R1   & R10   & R20   & mAP & R1   & R10   & R20   & mAP & R1 & R10   & R20   & mAP   \\ 
			\hline                   
			HOG \cite{r23} \iffalse2005\fi  &2.76 &18.25 &31.91 &4.24 &3.82 &22.77 &37.63 &2.16 &3.22 &24.68 &44.52 &7.25 &4.75 &29.06 &49.38 &3.51    \\  
			LOMO \cite{r24} \iffalse2007\fi &3.64 &23.18 &37.28 &4.53 &4.70 &28.22 &43.05 &2.28 &5.75 &34.35 &54.90 &10.19 &7.36 &40.38 &60.33 &5.64   \\   
			One-stream Network \iffalse2013\fi \cite{r8} &12.04 &49.68 &66.74 &13.67 &16.26 &58.14 &75.05 &8.59 &16.94 &63.55 &82.10 &22.95 &22.62 &71.74 &87.82 &15.04  \\
			Two-stream Network \cite{r8} &11.65 &47.99 &65.50 &12.85 &16.33 &58.35 &74.46 &8.03 &15.60 &61.18 &81.02 &21.49 &22.49 &72.22 &88.61 &13.92  \\
			Zero-padding \cite{r8} &14.80 &54.12 &71.33 &15.95 &19.13 &61.40 &78.41 &10.89 &20.58 &68.38 &85.79 &26.92 &24.43 &75.86 &91.32 &18.64  \\
			MLBP \cite{r25} \iffalse2017\fi & 2.1  &16.2  & 28.3  & 3.9    &-&-&-&-   &-&-&-&-  &-&-&-&-   \\  
			TONE\cite{r12} \iffalse2018\fi & 12.5  & 50.7 & 68.6  & 14.4   &-&-&-&- &-&-&-&- &-&-&-&- \\
			HCML\cite{r12} \iffalse2018\fi & 14.3  & 53.2 & 69.2  & 16.2   &-&-&-&- &-&-&-&- &-&-&-&-   \\
			BDTR \cite{r17} \iffalse2018\fi &17.0  &55.4 &72.0  & 19.7     &-&-&-&-  &26.8 &73.2 &87.6 &37.6  &-&-&-&-  \\
			cmGAN \cite{r7} \iffalse2018\fi & 27.0  & 67.5 & 80.6  & 27.8   &31.7 &77.2 &89.2 &42.4  &- &- &- &-  &- &- &- &- \\
			GSM \cite{r26} \iffalse2019\fi &5.29 &33.71 &52.95 &8.00 &6.19 &37.15 &55.66 &4.38 &9.46 &48.98 &72.06 &15.57 &11.36 &51.34 &73.41 &9.03    \\
			D-HSME \cite{r13}\iffalse2019\fi & 20.7  & 62.8 & 78.0  & 23.2   &-&-&-&-  &-&-&-&- &-&-&-&-   \\
			IPVT+MSR\cite{r29} \iffalse2019\fi& 23.2 & 51.2 & 61.7  & 22.5   &-&-&-&-  &-&-&-&- &-&-&-&- \\
			D$^2$RL \cite{r28}\iffalse2019\fi  & 28.9  & 70.6 & 82.4  & 29.2 &-&-&-&- &-&-&-&- &-&-&-&-   \\
			DBT \cite{r16} \iffalse2019\fi &29.05 &74.71 &87.16 &30.94 &35.40 &81.02 &91.85 &24.12 &32.74 &82.40 &93.35 &44.26 &40.41 &86.83 &96.27 &33.93 \\
			CC-S \cite{r2}  \iffalse2019\fi & 33.4  & 78.6 & 89.4  & 37.2  &- &- &- &-   &- &- &- &-  &- &- &- &-   \\
			CC-F \cite{r2} \iffalse2019\fi  & 35.1  & 77.6 & 88.9 & 37.4  &- &- &- &-   &- &- &- &-&- &- &- &-   \\
			DGD\_MSR\cite{r1} \iffalse2019\fi & 37.35  & 83.40 & 93.34  & 38.11 &43.86 &86.94 &95.68 &30.48  &39.64&89.29&97.66&50.88  &46.56 &93.57 &98.80 &40.08  \\
			AlignGAN\cite{r30} \iffalse2019\fi &42.4 &85.0 &93.7 &40.7 &51.5 &89.4 &95.7 &33.9 &45.9 &87.6 &94.4 &54.3 &57.1 &92.7 &97.4 &45.3 \\
			eBDTR \cite{r5} \iffalse2020\fi & 27.8  & 67.3 & 81.3  & 28.4 &-&-&-&-  &32.4 &77.4 &89.6 &42.4  &-&-&-&-  \\
			Hi-CMD\cite{r11} \iffalse2020\fi &34.94 &77.58 &-&35.94  &- &- &- &-  &- &- &- &- &- &- &- &-  \\
			EDFL \cite{r9} \iffalse2020\fi &36.9 &84.5 &93.2 &40.7 &- &- &- &-  &- &- &- &- &- &- &- &-  \\
			%	DPMBN \cite{r32} &37.02&79.46& 89.87 &40.28 &- &- &- &- &44.47 &87.12 &95.24 &54.51 &-& -& -& -\\
			CMPG \cite{r8} \iffalse2020\fi &38.1 &80.7 &89.9 &36.9 &45.1 &85.7 &93.8 &29.5 &43.8 &86.2 &94.2 &52.9 &52.7 &91.1 &96.4 &42.7   \\
			ECMC \cite{r76} \iffalse2020\fi &30.26 &75.59 &88.13 &33.38 &- &- &- &- &- &- &- &- &- &- &- &-   \\
			BEAT \cite{r77} \iffalse2020\fi &38.57 &76.64 &86.39 &38.61 &44.71 &69.82 &77.87 &32.20 &- &- &- &- &- &- &- &-   \\	
			%CDP+DHSM \cite{r79} \iffalse2020\fi &38.0 &82.3 &91.7 &38.4 &- &- &- &- &- &- &- &- &- &- &- &-   \\			
			HPILN\cite{r31}\iffalse2019\fi  &41.36 & 84.78  &94.51  &42.95  &47.56  &88.13  &95.98 & 36.08 & 45.77  &91.82  &98.46  &56.52  &53.05 & 93.71  &98.93  &47.48 \\
			XM \cite{r3} \iffalse2020\fi &49.9  &89.7 &95.9 &50.7 &- &- &- &-  &- &- &- &-  &- &- &- &-  \\
			TSGAN(w/o rerank) \cite{r80} \iffalse2020\fi &49.8  &87.3 &93.8 &47.4 &56.1 &90.2 &96.3 &38.5  &50.4 &90.8 &96.8 &63.1  &59.3 &91.2 &97.8 &50.2  \\
			TSGAN(w/ rerank) \cite{r80} \iffalse2020\fi &58.9  &87.8 &94.1 &55.1 &55.9 &91.2 &96.6 &39.7  &62.1 &90.8 &96.4 &71.3  &59.7 &91.8 &97.9 &50.9  \\
			ABP \cite{r67} \iffalse2020\fi &51.56  &75.65 &81.69 &32.50 &- &- &- &-  &- &- &- &-  &- &- &- &- \\
			%DDAG \cite{r78} \iffalse2020\fi &54.75  &90.39 &95.81 &53.02 &61.02 &94.06 &98.41 &67.98 &- &- &- &- &- &- &- &-   \\
			%MCEL \cite{r87} \iffalse2020\fi &51.64 &87.25 &94.44 &50.11 & - &- &- &- &57.35 &93.02 &97.48& 64.79& - &- &- &- \\
			HATML \cite{r71} \iffalse2020\fi &55.29  &92.41 &97.36 &53.89 &- &- &- &-  &62.10 &95.75 &99.20 &69.37  &- &- &- &- \\
			HC \cite{r4} \iffalse2020\fi &56.96 &91.50 &96.82 &54.95& 62.09 &93.74 &97.85 &48.02 &59.74 &92.07 &96.22& 64.91& 69.76 &95.85 &98.90 &57.81  \\
			DG-VAE \cite{r74} \iffalse2020\fi &59.49 &93.77 &- &58.46& - &- &- &- &- &- &-& -& - &- &- &-  \\
			our  &\textbf{62.56} &\textbf{93.85} & \textbf{97.63} &\textbf{60.57} &\textbf{68.09}  &\textbf{95.73} &\textbf{98.30} &\textbf{54.33} &\textbf{65.06} &\textbf{95.17} &\textbf{98.17}&\textbf{73.86} &\textbf{69.81}  &\textbf{96.86}  &\textbf{99.30} &\textbf{64.84 }  \\ 
			\hline 
		\end{tabular}
	}
\end{table*}

\begin{table*}[!t]
	\renewcommand{\arraystretch}{1.3}
	\caption{Comparison with the state-of-the-art models on RegDB dataset.}
	\label{tab3}
	\centering
	\begin{tabular}{|c|cccc|cccc|}
		\hline	
		-	  &  \multicolumn{4}{c}{Visible to Thermal}  & \multicolumn{4}{|c|}{Thermal to Visible} \\
		\hline	
		Methods	  & R1   & R10   & R20   & mAP & R1   & R10   & R20   & mAP  \\ 
		\hline                   
		HOG \cite{r23}  &13.5 &33.2 &43.7 &10.3 &-& -&- &   \\  
		MLBP \cite{r25} & 2.0 &7.3 &10.9 &6.8   &-& -&- & \\  
		LOMO \cite{r24} &0.9 &2.5 &4.1 &2.3     &-& -&- & \\   
		GSM \cite{r26}  &17.3 &34.5 &45.3 &15.1  &-& -&- &       \\
		One-stream Network \cite{r8} &13.1 &33.0 &42.5 &14.0 &-& -&- &- \\
		Two-stream Network \cite{r8} &12.4 &30.4 &41.0 &13.4 &-& -&- &- \\
		Zero-padding \cite{r8} &17.8 &34.2 &44.4 &18.9 &16.63 &34.68 &44.25 &17.82 \\
		TONE\cite{r12}  &16.9 &34.0 &44.1 &14.9 &13.86 &30.08 &40.05 &16.98 \\
		HCML\cite{r12}  &24.4 &47.5 &56.8 &20.8   &21.70 &45.02 &55.58 &22.24  \\
		D-HSME \cite{r13} &50.85 &73.36 &81.66 &47.00 &50.15 &72.40 &81.07 &46.16  \\
		BDTR \cite{r17} &33.5 &58.4 &67.5 &31.8   & 32.92 &58.46  &68.43  &31.96   \\
		eBDTR \cite{r5} &31.8 &56.1 &66.8 &33.2  &34.21  &58.74  &68.64  &32.49   \\
		IPVT+MSR\cite{r29}  &58.76& 85.75&90.27 &47.85  &-& -&- &-   \\
		D$^2$RL \cite{r28} &43.4 &66.1 &76.3 &44.1  &-& -&- &-   \\
		DBT \cite{r16} &38.64 &60.18 &69.81 &38.08 &-& -&- &-     \\
		CC-S \cite{r2} &53.1 &72.3 &80.2 &53.5  &-& -&- &-   \\
		CC-F \cite{r2}  &30.26 &75.59 &88.13 &60.0 &-& -&- &-    \\
		ECMC \cite{r76} &39.75 &61.26 &70.10 &40.79 &-& -&- &-    \\
		BEAT \cite{r77} &67.45 &- &- &66.51 &66.48& -&- &67.31    \\
		Hi-CMD\cite{r11} &70.93 &86.39 &-&66.04 &-& -&- &-   \\
		EDFL \cite{r9} &52.58 &72.10 &81.47 &52.98  &-& -&- &-  \\
		DGD\_MSR\cite{r1} &48.43 &70.32 &79.59 &48.67  &-& -&- &-   \\
		AlignGAN\cite{r30} &56.3 &53.4 &57.9 &53.6 &  &  &  &   \\
		XM \cite{r3} &62.21 &83.13 &91.72 &60.18 &57.9  &-  &-  & 53.6 \\ 
		ABP \cite{r67} &56.35 &80.87 &87.96 &48.58 &54.03  &78.69 &85.83  & 47.60 \\ 
	    HATML \cite{r71} \iffalse2020\fi &71.83 &87.16 &92.16 &67.56 &70.02  &86.45 &91.61  & 66.30 \\
	    %MCEL \cite{r87} \iffalse2020\fi &72.12 &88.07 &93.07 &69.09 &72.12  &88.07 &93.07  & 68.57 \\
	    DG-VAE \cite{r74} \iffalse2020\fi &72.97 &86.89 &- &71.78 &-  &- &-  & - \\ 
		our 	&\textbf{76.10}	&\textbf{88.86}	&\textbf{92.41}	&\textbf{74.39} &\textbf{72.18}&\textbf{87.06}  &\textbf{92.38} &\textbf{71.04} \\	
		\hline
	\end{tabular}
\end{table*}

The effectiveness of each component in our proposed model is studied in this subsection. As shown in Table \ref{tab1}, different versions of our proposed model are designed for comparisons. Specifically, the modality-invariant relation features, multi-level features and PCB-based part-aligned block are first removed from the proposed model as the Baseline model (\emph{i.e.,} the first line of Table \ref{tab1}). Then, the multi-level features are represented as `ML' and the PCB-based part-aligned block is represented as `P'. `RF' denotes that the modality-invariant relation features are employed for cross-modality person Re-ID.  `RF-only' represents that only modality-invariant relation features are employed, while modality-shared appearance features are not extracted for cross-modality person Re-ID. Finally, the proposed cross-modality quadruplet loss is represented as `CQ'. It should be noted that these models are trained by using the joint loss of classification loss, bi-directional dual-constrained top-ranking loss \cite{r5} and single-modality triplet loss, except for the last model in Table \ref{tab1}, in which the proposed cross-modality quadruplet loss, instead of the bi-directional dual-constrained top-ranking loss, is adopted for training.

As shown in the second and third rows of Table \ref{tab1}, the performance of cross-modality person Re-ID are boosted by employing the multi-level features and the PCB-based algorithm. This results from that both multi-level features and PCB-based algorithm can enhance the discriminability of the extracted modality-shared features. Furthermore, the results shown in the fourth row indicate that only employing the modality-shared appearance features for cross-modality person Re-ID cannot achieve the optimal results. This may be due to the fact that using modality-shared appearance features only cannot capture enough discriminative information for identifying different persons since the modality differences between RGB and IR images are large. Meanwhile, the results in the fifth row of Table \ref{tab1} indicate that employing modality-invariant relation features only also obtains suboptimal results. This may result from the fact the modality-shared appearance features also contain lots of valuable information for identifying different persons. The modality-invariant relation features can be employed as important complementary information for modality-shared appearance features but cannot fully replace the modality-shared appearance features. Furthermore, as shown in the sixth row of Table \ref{tab1}, the performance of cross-modality person Re-ID is significantly boosted by simultaneously capturing the modality-shared appearance features and modality-invariant relation features. This means that, by virtue of modality-shared appearance features and modality-invariant relation features, our proposed MTMFE sub-network can extract more modality-invariant and discriminative modality-shared features to reduce cross-modality variations and intra-modality variations, thus boosting the cross-modality person Re-ID.  As shown in the last row of Table \ref{tab1}, the performance of our model is further improved by employing the proposed cross-modality quadruplet loss. This results from the fact that the proposed cross-modality quadruplet loss provides stronger constraints to force our model to learn more discriminative features in the training process.

\subsection{Comparison with State-of-the-Art Methods}

In this subsection, the proposed model is compared with some of the State-Of-The-Art (SOTA) methods, including: HOG \cite{r23}, MLBP \cite{r25}, LOMO \cite{r24}, GSM \cite{r26}, One-stream Network\cite{r8}, Two-stream Network \cite{r8}, Zero-padding \cite{r8}, TONE\cite{r12}, HCML\cite{r12}, D-HSME \cite{r13}, BDTR \cite{r17}, eBDTR \cite{r5}, cmGAN \cite{r7}, IPVT+MSR\cite{r29}, D$^2$RL \cite{r28}, DBT \cite{r16}, CC \cite{r2}, Hi-CMD\cite{r11}, EDFL \cite{r9},  DGD\_MSR\cite{r1}, CMPG \cite{r8}, AlignGAN\cite{r30}, TSGAN\cite{r80}, BEAT \cite{r77}, ECMC \cite{r76}, HPILN\cite{r31}, XM \cite{r3}, HC \cite{r4}, ABP \cite{r67} and HATML \cite{r71}. % MCEL \cite{r87}

The quantitative results of these SOTA models on SYSU-MM01 \cite{r15} dataset are shown in Table \ref{tab2}. It can be seen that the proposed model outperforms other SOTA methods by a large margin. Specifically, in all-search mode, the proposed model surpasses SOAT models by 3.07\% on Rank-1 accuracy and 2.11\% on mAP in the single-shot setting. Meanwhile, the multi-shot setting exhibits a similar phenomenon. This indicates that the proposed model can extract highly discriminative person features by jointly capturing modality-shared appearance features and modality-invariant relation features. Furthermore,  our proposed model can obtain higher recall  values than other SOTA models when the gallery size increases. Meanwhile, the proposed model also obtains the best performance in the indoor-search mode, which further demonstrates the effectiveness and robustness of our proposed model.

The quantitative results of these SOTA models on RegDB dataset are shown in Table \ref{tab3}. It can be seen that, for both the visible-to-thermal mode and the thermal-to-visible mode, the proposed model surpasses others by a large margin. Meanwhile, the proposed model obtains similar mAP values in the visible-to-thermal mode and thermal-to-visible mode. This also indicates that, with the collaboration of modality-shared appearance features, modality-invariant relation features and cross-modality quadruplet loss, the proposed model can significantly reduce the cross-modality variations and intra-modality variations for cross-modality person Re-ID. As a result, more discriminative modality-shared features are extracted to boost the performance of cross-modality person Re-ID.

\section{Conclusion} \label{sec::con}

A novel cross-modality person Re-ID model has been presented in this paper. Specifically, on the top of modality-shared appearance features, our proposed MTMFE sub-network enhances the discriminability of the extracted modality-shared features by further extracting modality-invariant relation features. By virtue of the extracted modality-invariant relation features, the cross-modality variations as well as the intra-modality variations are significantly reduced. As a result, the performance of cross-modality person Re-ID is greatly improved. After that, the cross-modality variations are further reduced by employing the proposed cross-modality quadruplet loss. With the collaboration of modality-shared appearance features, modality-invariant relation features and cross-modality quadruplet loss, the proposed model achieves new state-of-the-art experimental results on several benchmarks, which validates the superiorities of our model over others.

%To reduce the large cross-modality variants and obtain more discriminative modality-shared features for cross-modality person Re-ID, a novel end-to-end trainable network is presented in this paper. In the proposed network, the cross-modality variants are significantly reduced and the discriminability of the extracted modality-shared features is greatly enhanced by simultaneously extracting shared appearance features and shared relation features through the proposed MTMFE sub-network. As a result, the performance of cross-modality person Re-ID is greatly improved. Meanwhile, the cross-modality variants are further reduced by employing the proposed cross-modality quadruplet loss. The proposed model achieves new state-of-the-art experimental results on several benchmarks, which validates the superiorities of our model over others.

\section*{Acknowledgment}
This work is supported by the National Natural Science Foundation of China under Grant No. 61773301 and 61876140, and the China Postdoctoral Support Scheme for Innovative Talents under Grant No. BX20180236.

\ifCLASSOPTIONcaptionsoff
\newpage
\fi

\bibliographystyle{IEEEtran}
\bibliography{IEEEabrv,work20}

\begin{thebibliography}{10}
\providecommand{\url}[1]{#1}
\csname url@samestyle\endcsname
\providecommand{\newblock}{\relax}
\providecommand{\bibinfo}[2]{#2}
\providecommand{\BIBentrySTDinterwordspacing}{\spaceskip=0pt\relax}
\providecommand{\BIBentryALTinterwordstretchfactor}{4}
\providecommand{\BIBentryALTinterwordspacing}{\spaceskip=\fontdimen2\font plus
\BIBentryALTinterwordstretchfactor\fontdimen3\font minus
  \fontdimen4\font\relax}
\providecommand{\BIBforeignlanguage}[2]{{%
\expandafter\ifx\csname l@#1\endcsname\relax
\typeout{** WARNING: IEEEtran.bst: No hyphenation pattern has been}%
\typeout{** loaded for the language `#1'. Using the pattern for}%
\typeout{** the default language instead.}%
\else
\language=\csname l@#1\endcsname
\fi
#2}}
\providecommand{\BIBdecl}{\relax}
\BIBdecl

\bibitem{r18}
X.~Wang, ``Intelligent multi-camera video surveillance: A review,''
  \emph{Pattern Recognition Letters}, vol.~34, pp. 3--19, 2013.

\bibitem{r19}
N.~McLaughlin, J.~M. del Rincon, and P.~Miller, ``Video person
  re-identification for wide area tracking based on recurrent neural
  networks,'' \emph{IEEE Transactions on Circuits and Systems for Video
  Technology}, vol.~29, pp. 2613--2626.

\bibitem{r36}
L.~Zheng, Y.~Huang, H.~Lu, and Y.~Yang, ``Pose-invariant embedding for deep
  person re-identification,'' \emph{IEEE Transactions on Image Processing},
  vol.~28, no.~9, pp. 4500--4509, 2019.

\bibitem{r37}
Z.~Feng, J.~Lai, and X.~Xie, ``Learning view-specific deep networks for person
  re-identification,'' \emph{IEEE Transactions on Image Processing}, vol.~27,
  no.~7, pp. 3472--3483, 2018.

\bibitem{r38}
Z.~Zheng, L.~Zheng, and Y.~Yang, ``Pedestrian alignment network for large-scale
  person re-identification,'' \emph{IEEE Transactions on Circuits and Systems
  for Video Technology}, vol.~29, no.~10, pp. 3037--3045, 2018.

\bibitem{r41}
B.~Nguyen and B.~De~Baets, ``Kernel distance metric learning using pairwise
  constraints for person re-identification,'' \emph{IEEE Transactions on Image
  Processing}, vol.~28, no.~2, pp. 589--600, 2018.

\bibitem{r42}
Y.-J. Cho and K.-J. Yoon, ``Pamm: Pose-aware multi-shot matching for improving
  person re-identification,'' \emph{IEEE Transactions on Image Processing},
  vol.~27, no.~8, pp. 3739--3752, 2018.

\bibitem{r43}
A.~Hermans, L.~Beyer, and B.~Leibe, ``In defense of the triplet loss for person
  re-identification,'' \emph{ArXiv}, vol. abs/1703.07737, 2017.

\bibitem{r1}
Z.~Feng, J.~Lai, and X.~Xie, ``Learning modality-specific representations for
  visible-infrared person re-identification,'' \emph{IEEE Transactions on Image
  Processing}, vol.~29, pp. 579--590, 2019.

\bibitem{r2}
S.~Zhang, Y.~Yang, P.~Wang, X.~Zhang, and Y.~Zhang, ``Attend to the difference:
  Cross-modality person re-identification via contrastive correlation,''
  \emph{arXiv preprint arXiv:1910.11656}, 2019.

\bibitem{r3}
D.~Li, X.~Wei, X.~Hong, and Y.~Gong, ``Infrared-visible cross-modal person
  re-identification with an {X} modality,'' in \emph{Proceedings of the AAAI
  Conference on Artificial Intelligence}, 2020, pp. 4610--4617.

\bibitem{r4}
Y.~Zhu, Z.~Yang, L.-C. Wang, S.~Zhao, X.~Hu, and D.~Tao, ``Hetero-center loss
  for cross-modality person re-identification,'' \emph{Neurocomputing}, vol.
  386, pp. 97--109, 2020.

\bibitem{r5}
M.~Ye, X.~Lan, Z.~Wang, and P.~C. Yuen, ``Bi-directional center-constrained
  top-ranking for visible thermal person re-identification,'' \emph{IEEE
  Transactions on Information Forensics and Security}, vol.~15, pp. 407--419,
  2020.

\bibitem{r6}
E.~Basaran, M.~G{\"o}kmen, and M.~E. Kamasak, ``An efficient framework for
  visible-infrared cross modality person re-identification,'' \emph{Signal
  Processing: Image Communication}, p. 115933, 2020.

\bibitem{r7}
P.~Dai, R.~Ji, H.~Wang, Q.~Wu, and Y.~Huang, ``Cross-modality person
  re-identification with generative adversarial training,'' in
  \emph{Proceedings of the International Joint Conference on Artificial
  Intelligence}, 2018, pp. 677--683.

\bibitem{r9}
H.~Liu, J.~Cheng, W.~Wang, Y.~Su, and H.~Bai, ``Enhancing the discriminative
  feature learning for visible-thermal cross-modality person
  re-identification,'' \emph{Neurocomputing}, vol. 398, pp. 11--19, 2020.

\bibitem{r17}
M.~Ye, Z.~Wang, X.~Lan, and P.~C. Yuen, ``Visible thermal person
  re-identification via dual-constrained top-ranking,'' in \emph{Proceedings of
  the International Joint Conference on Artificial Intelligence}, 2018, pp.
  1092--1099.

\bibitem{r16}
N.~Tekeli and A.~B. Can, ``Distance based training for cross-modality person
  re-identification,'' in \emph{Proceedings of the IEEE International
  Conference on Computer Vision Workshop}, 2019, pp. 4540--4549.

\bibitem{r13}
Y.~Hao, N.~Wang, L.~Jie, and X.~Gao, ``{HSME}: Hypersphere manifold embedding
  for visible thermal person re-identification,'' in \emph{Proceedings of the
  AAAI Conference on Artificial Intelligence}, 2019, pp. 8385--8392.

\bibitem{r12}
M.~Ye, X.~Lan, J.~Li, and P.~C. Yuen, ``Hierarchical discriminative learning
  for visible thermal person re-identification,'' in \emph{Proceedings of the
  AAAI Conference on Artificial Intelligence}, 2018, pp. 7501--7508.

\bibitem{r86}
J.~Zhou, B.~Su, and Y.~Wu, ``Easy identification from better constraints:
  Multi-shot person re-identification from reference constraints,'' in
  \emph{Proceedings of the IEEE Conference on Computer Vision and Pattern
  Recognition}, 2018, pp. 5373--5381.

\bibitem{r84}
Z.~Qiu, T.~Yao, and T.~Mei, ``Learning spatio-temporal representation with
  pseudo-3d residual networks,'' in \emph{Proceedings of the IEEE International
  Conference on Computer Vision}, 2017, pp. 5533--5541.

\bibitem{r85}
D.~Tran, H.~Wang, L.~Torresani, J.~Ray, Y.~LeCun, and M.~Paluri, ``A closer
  look at spatiotemporal convolutions for action recognition,'' in
  \emph{Proceedings of the IEEE Conference on Computer Vision and Pattern
  Recognition}, 2018, pp. 6450--6459.

\bibitem{r53}
S.~Ji, W.~Xu, M.~Yang, and K.~Yu, ``{3D} convolutional neural networks for
  human action recognition,'' \emph{IEEE Transactions on Pattern Analysis and
  Machine Intelligence}, vol.~35, no.~1, pp. 221--231, 2013.

\bibitem{r54}
M.~Sabokrou, M.~Fayyaz, M.~Fathy, and R.~Klette, ``Deep-cascade: Cascading {3D}
  deep neural networks for fast anomaly detection and localization in crowded
  scenes,'' \emph{IEEE Transactions on Image Processing}, vol.~26, no.~4, pp.
  1992--2004, 2017.

\bibitem{r52}
D.~Tran, L.~D. Bourdev, R.~Fergus, L.~Torresani, and M.~Paluri, ``Learning
  spatiotemporal features with {3D} convolutional networks,'' in
  \emph{Proceedings of the IEEE International Conference on Computer Vision},
  2015, pp. 4489--4497.

\bibitem{r10}
Y.~Lu, Y.~Wu, B.~Liu, T.~Zhang, B.~Li, Q.~Chu, and N.~Yu, ``Cross-modality
  person re-identification with shared-specific feature transfer,'' in
  \emph{Proceedings of the IEEE Conference on Computer Vision and Pattern
  Recognition}, 2020, pp. 13\,379--13\,389.

\bibitem{r31}
Y.-B. Zhao, J.-W. Lin, Q.~Xuan, and X.~Xi, ``Hpiln: a feature learning
  framework for cross-modality person re-identification,'' \emph{IET Image
  Processing}, vol.~13, no.~14, pp. 2897--2904, 2019.

\bibitem{r34}
S.~Karanam, M.~Gou, Z.~Wu, A.~Rates-Borras, O.~I. Camps, and R.~J. Radke, ``A
  systematic evaluation and benchmark for person re-identification: Features,
  metrics, and datasets,'' \emph{IEEE Transactions on Pattern Analysis and
  Machine Intelligence}, vol.~41, pp. 523--536, 2019.

\bibitem{r35}
M.~A. Saghafi, A.~Hussain, H.~B. Zaman, and M.~H.~M. Saad, ``Review of person
  re-identification techniques,'' \emph{IET Computer Vision}, vol.~8, no.~6,
  pp. 455--474, 2014.

\bibitem{r39}
Z.~Feng, J.~Lai, and X.~Xie, ``Learning view-specific deep networks for person
  re-identification,'' \emph{IEEE Transactions on Image Processing}, vol.~27,
  no.~7, pp. 3472--3483, 2018.

\bibitem{r46}
H.~Yao, S.~Zhang, R.~Hong, Y.~Zhang, C.~Xu, and Q.~Tian, ``Deep representation
  learning with part loss for person re-identification,'' \emph{IEEE
  Transactions on Image Processing}, vol.~28, no.~6, pp. 2860--2871, 2019.

\bibitem{r44}
W.~Chen, X.~Chen, J.~Zhang, and K.~Huang, ``Beyond triplet loss: A deep
  quadruplet network for person re-identification,'' in \emph{Proceedings of
  the IEEE Conference on Computer Vision and Pattern Recognition}, 2017, pp.
  1320--1329.

\bibitem{r45}
J.~Chen, Z.~Zhang, and Y.~Wang, ``Relevance metric learning for person
  re-identification by exploiting listwise similarities,'' \emph{IEEE
  Transactions on Image Processing}, vol.~24, no.~12, pp. 4741--4755, 2015.

\bibitem{r14}
M.~Ye, X.~Lan, and Q.~Leng, ``Modality-aware collaborative learning for visible
  thermal person re-identification,'' in \emph{Proceedings of the ACM
  International Conference on Multimedia}, 2019, pp. 347--355.

\bibitem{r8}
Y.~Yang, T.~Zhang, J.~Cheng, Z.~Hou, P.~Tiwari, H.~M. Pandey \emph{et~al.},
  ``Cross-modality paired-images generation and augmentation for {RGB-Infrared}
  person re-identification,'' \emph{Neural Networks}, vol. 128, pp. 294--304,
  2020.

\bibitem{r75}
A.~Krizhevsky, I.~Sutskever, and G.~E. Hinton, ``Imagenet classification with
  deep convolutional neural networks,'' in \emph{The proceedings of the Neural
  Information Processing Systems Conference}, 2012, pp. 1097--1105.

\bibitem{r30}
G.~Wang, T.~Zhang, J.~Cheng, S.~Liu, Y.~Yang, and Z.~Hou, ``{RGB-Infrared}
  cross-modality person re-identification via joint pixel and feature
  alignment,'' in \emph{Proceedings of the IEEE International Conference on
  Computer Vision}, 2019, pp. 3622--3631.

\bibitem{r49}
Z.~Sun, X.~Wang, Q.~Zhang, and J.~Jiang, ``Real-time video saliency prediction
  via {3D} residual convolutional neural network,'' \emph{IEEE Access}, vol.~7,
  pp. 147\,743--147\,754, 2019.

\bibitem{r64}
T.~M. Lee, J.-C. Yoon, and I.-K. Lee, ``Motion sickness prediction in
  stereoscopic videos using {3D} convolutional neural networks,'' \emph{IEEE
  Transactions on Visualization and Computer Graphics}, vol.~25, no.~5, pp.
  1919--1927, 2019.

\bibitem{r50}
S.~Ji, W.~Xu, M.~Yang, and K.~Yu, ``{3D} convolutional neural networks for
  human action recognition,'' \emph{IEEE Transactions on Pattern Analysis and
  Machine Intelligence}, vol.~35, no.~1, pp. 221--231, 2012.

\bibitem{r66}
Y.~Xiang, H.~Liu, S.~Wang, L.~Ma, X.~Xiong, C.~Xu, and D.~Shao, ``Segmentation
  method of multiple sclerosis lesions based on 3d-cnn networks,'' \emph{IET
  Image Processing}, pp. 1806--1812, 2020.

\bibitem{r65}
Q.~Dou, H.~Chen, L.~Yu, L.~Zhao, J.~Qin, D.~Wang, V.~C. Mok, L.~Shi, and P.-A.
  Heng, ``Automatic detection of cerebral microbleeds from mr images via 3d
  convolutional neural networks,'' \emph{IEEE Transactions on Medical Imaging},
  vol.~35, no.~5, pp. 1182--1195, 2016.

\bibitem{r20}
K.~He, X.~Zhang, S.~Ren, and J.~Sun, ``Deep residual learning for image
  recognition,'' in \emph{Proceedings of the IEEE Conference on Computer Vision
  and Pattern Recognition}, 2015, pp. 770--778.

\bibitem{r61}
Y.~Sun, L.~Zheng, Y.~Yang, Q.~Tian, and S.~Wang, ``Beyond part models: Person
  retrieval with refined part pooling and a strong convolutional baseline,'' in
  \emph{Proceedings of the European Conference on Computer Vision}, 2018, pp.
  501--518.

\bibitem{r15}
A.~Wu, W.-S. Zheng, H.-X. Yu, S.~Gong, and J.-H. Lai, ``{RGB-Infrared}
  cross-modality person re-identification,'' in \emph{Proceedings of the IEEE
  International Conference on Computer Vision}, 2017, pp. 5390--5399.

\bibitem{r57}
D.~T. Nguyen, H.~G. Hong, K.~W. Kim, and K.~R. Park, ``Person recognition
  system based on a combination of body images from visible light and thermal
  cameras,'' \emph{Sensors}, vol.~17, no.~3, p. 605, 2017.

\bibitem{r67}
Z.~Wei, X.~Yang, N.~Wang, B.~Song, and X.~Gao, ``{ABP}: Adaptive body partition
  model for visible infrared person re-identification,'' in \emph{Proceedings
  of the IEEE International Conference on Multimedia and Expo}, 2020, pp. 1--6.

\bibitem{r59}
A.~Paszke, S.~Gross, F.~Massa, A.~Lerer, J.~Bradbury, G.~Chanan, T.~Killeen,
  Z.~Lin, N.~Gimelshein, L.~Antiga, A.~Desmaison, A.~K{\"o}pf, E.~Yang,
  Z.~DeVito, M.~Raison, A.~Tejani, S.~Chilamkurthy, B.~Steiner, L.~Fang,
  J.~Bai, and S.~Chintala, ``Pytorch: An imperative style, high-performance
  deep learning library,'' in \emph{Proceedings of the Neural Information
  Processing Systems}, 2019, pp. 8026--8037.

\bibitem{r22}
J.~Deng, W.~Dong, R.~Socher, L.~Li, K.~Li, and F.~Li, ``{ImageNet}: A
  large-scale hierarchical image database,'' in \emph{Proceedings of the IEEE
  Conference on Computer Vision and Pattern Recognition}, 2009, pp. 248--255.

\bibitem{r60}
X.~Glorot and Y.~Bengio, ``Understanding the difficulty of training deep
  feedforward neural networks,'' in \emph{Proceedings of the International
  Conference on Artificial Intelligence and Statistics}, 2010, pp. 249--256.

\bibitem{r23}
N.~Dalal and B.~Triggs, ``Histograms of oriented gradients for human
  detection,'' in \emph{Proceedings of the IEEE Computer Society Conference on
  Computer Vision and Pattern Recognition}, vol.~1, 2005, pp. 886--893.

\bibitem{r24}
S.~Liao, X.~Zhu, Z.~Lei, L.~Zhang, and S.~Z. Li, ``Learning multi-scale block
  local binary patterns for face recognition,'' in \emph{Proceedings of the
  International Conference on Biometrics}, 2007, pp. 828--837.

\bibitem{r25}
S.~Liao and S.~Z. Li, ``Efficient {PSD} constrained asymmetric metric learning
  for person re-identification,'' in \emph{Proceedings of the IEEE
  International Conference on Computer Vision}, 2015, pp. 3685--3693.

\bibitem{r26}
L.~Lin, G.~Wang, W.~Zuo, X.~Feng, and L.~Zhang, ``Cross-domain visual matching
  via generalized similarity measure and feature learning,'' \emph{IEEE
  Transactions on Pattern Analysis and Machine Intelligence}, vol.~39, pp.
  1089--1102, 2017.

\bibitem{r29}
J.~K. Kang, T.~M. Hoang, and K.~R. Park, ``Person re-identification between
  visible and thermal camera images based on deep residual {CNN} using single
  input,'' \emph{IEEE Access}, vol.~7, pp. 57\,972--57\,984, 2019.

\bibitem{r28}
Z.~Wang, Z.~Wang, Y.~Zheng, Y.-Y. Chuang, and S.~Satoh, ``Learning to reduce
  dual-level discrepancy for infrared-visible person re-identification,'' in
  \emph{Proceedings of theIEEE Conference on Computer Vision and Pattern
  Recognition}, 2019, pp. 618--626.

\bibitem{r11}
S.~Choi, S.~Lee, Y.~Kim, T.~Kim, and C.~Kim, ``{Hi-CMD}: Hierarchical
  cross-modality disentanglement for visible-infrared person
  re-identification,'' in \emph{Proceedings of the IEEE Conference on Computer
  Vision and Pattern Recognition}, 2020, pp. 10\,257--10\,266.

\bibitem{r76}
D.~Cheng, X.~Li, M.~Qi, X.~Liu, C.~Chen, and D.~Niu, ``Exploring cross-modality
  commonalities via dual-stream multi-branch network for infrared-visible
  person re-identification,'' \emph{IEEE Access}, vol.~8, pp. 12\,824--12\,834,
  2020.

\bibitem{r77}
H.~Ye, H.~Liu, F.~Meng, and X.~Li, ``Bi-directional exponential angular triplet
  loss for rgb-infrared person re-identification,'' \emph{IEEE Transactions on
  Image Processing}, vol.~30, pp. 1583--1595, 2020.

\bibitem{r80}
Z.~Zhang, S.~Jiang, C.~Huang, Y.~Li, and R.~Y. Da~Xu, ``{RGB-IR} cross-modality
  person reid based on teacher-student {GAN} model,'' \emph{arXiv preprint
  arXiv:2007.07452}, 2020.

\bibitem{r71}
M.~Ye, J.~Shen, and L.~Shao, ``Visible-infrared person re-identification via
  homogeneous augmented tri-modal learning,'' \emph{IEEE Transactions on
  Information Forensics and Security}, vol.~16, pp. 728--739, 2020.

\bibitem{r74}
N.~Pu, W.~Chen, Y.~Liu, E.~M. Bakker, and M.~S. Lew, ``Dual gaussian-based
  variational subspace disentanglement for visible-infrared person
  re-identification,'' in \emph{Proceedings of the ACM International Conference
  on Multimedia}, 2020, pp. 2149--2158.

\end{thebibliography}

\end{document}